\documentclass[10pt,twocolumn,letterpaper]{article}

\usepackage[pagenumbers]{cvpr} 

\usepackage{multirow} 
\usepackage{amssymb} 
\usepackage{pifont}

\usepackage[dvipsnames]{xcolor}

\definecolor{cvprblue}{rgb}{0.21,0.49,0.74}
\usepackage[pagebackref,breaklinks,colorlinks,citecolor=cvprblue]{hyperref}

\def\paperID{754} 
\def\confName{CVPR}
\def\confYear{2024}

\title{FocSAM: Delving Deeply into Focused Objects in Segmenting Anything}

\author{   
\normalsize You Huang$^{1}$, Zongyu Lan$^{1}$, Liujuan Cao$^{1}$\thanks{Corresponding author}, Xianming Lin$^{1}$, Shengchuan Zhang$^{1}$, Guannan Jiang$^{2}$, Rongrong Ji$^{1}$ \\
\normalsize $^1$ Key Laboratory of Multimedia Trusted Perception and Efficient Computing, \\
\normalsize Ministry of Education of China, Xiamen University \\
\normalsize $^2$ Intelligent Manufacturing Department, Contemporary Amperex Technology Co. Limited (CATL)
}

\begin{document}
\maketitle
\begin{abstract}
    The Segment Anything Model (SAM) marks a notable milestone in segmentation models, highlighted by its robust zero-shot capabilities and ability to handle diverse prompts. SAM follows a pipeline that separates interactive segmentation into image preprocessing through a large encoder and interactive inference via a lightweight decoder, ensuring efficient real-time performance. However, SAM faces stability issues in challenging samples upon this pipeline. These issues arise from two main factors. Firstly, the image preprocessing disables SAM to dynamically use image-level zoom-in strategies to refocus on the target object during interaction. Secondly, the lightweight decoder struggles to sufficiently integrate interactive information with image embeddings. To address these two limitations, we propose FocSAM with a pipeline redesigned on two pivotal aspects. 
    First, we propose Dynamic Window Multi-head Self-Attention (Dwin-MSA) to dynamically refocus SAM's image embeddings on the target object. Dwin-MSA localizes attention computations around the target object, enhancing object-related embeddings with minimal computational overhead.
    Second, we propose Pixel-wise Dynamic ReLU (P-DyReLU) to enable sufficient integration of interactive information from a few initial clicks that have significant impacts on the overall segmentation results. Experimentally, FocSAM augments SAM’s interactive segmentation performance to match the existing state-of-the-art method in segmentation quality, requiring only about $5.6\%$ of this method's inference time on CPUs. Code is available at \url{https://github.com/YouHuang67/focsam}.
\end{abstract}
\section{Introduction}

\begin{figure}[t!]  
\centering
\includegraphics[width=0.94\linewidth]{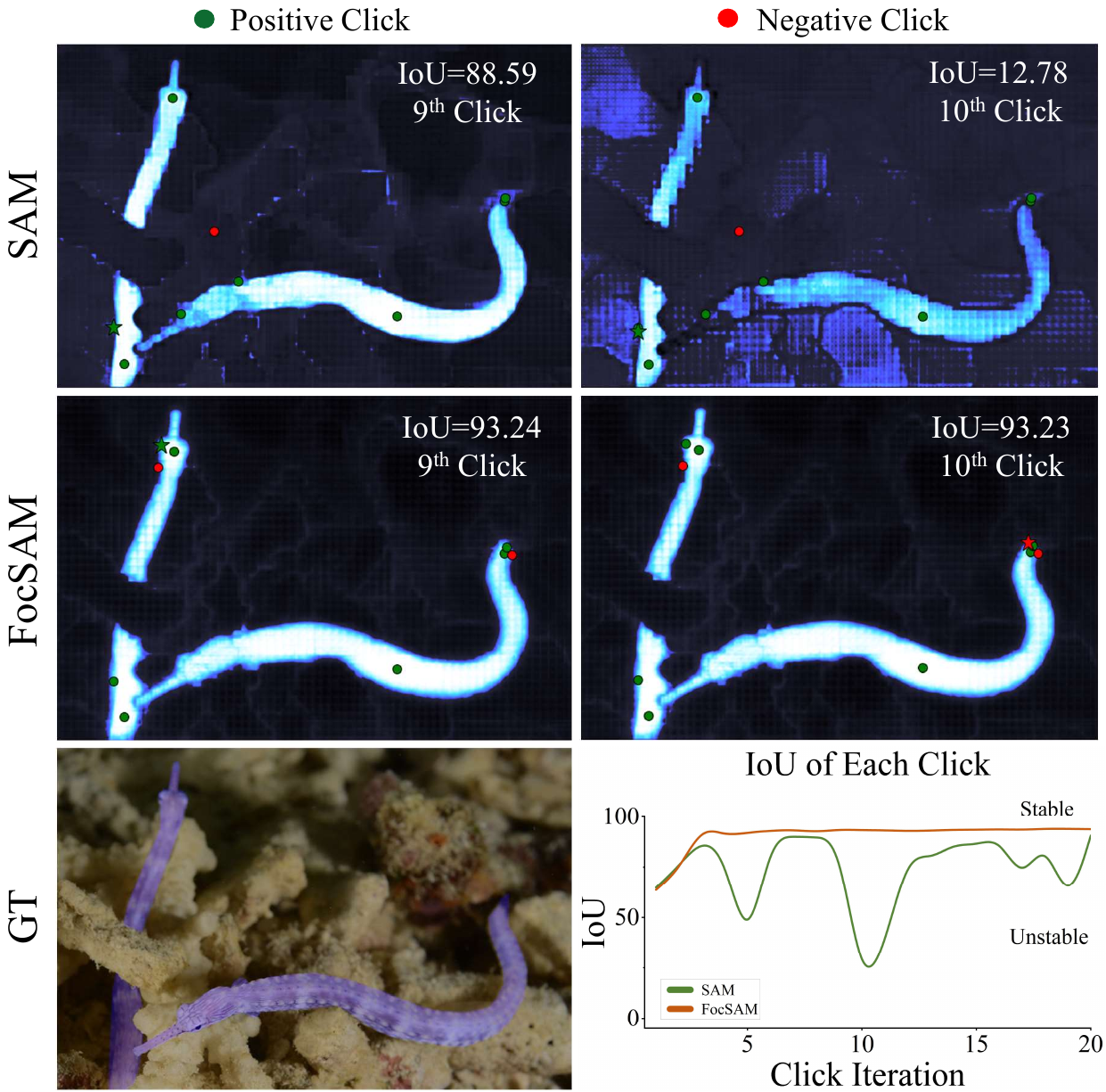}
\vspace{-0.7em}
\caption{
    Interactive segmentation stability on a challenging example. The bottom-left shows the example overlaid with GT (purple masks). 
    The top and middle rows illustrate the interactive segmentation of SAM and the proposed FocSAM, where each click is placed at the center of erroneously predicted regions and categorized as either positive (green) or negative (red).
    SAM's performance is unstable in this example (top row), where the $9$th click yields an IoU of $88.59$ (left) but a subsequent click significantly reduces the IoU to $12.78$ (right).
    In contrast, FocSAM (middle row) shows consistent performance. The plot (bottom-right) summarizes the trends of 20 clicks's segmentation, clearly contrasting SAM's IoU fluctuations with FocSAM's stable performance.
}
\label{figure:motivation}
\vspace{-.5em}
\end{figure}

\label{sec:intro}
Interactive segmentation~\cite{XiChen2022FocalClickTP,QinLiu2022SimpleClickII,huang2023interformer,kirillov2023segany} enhances the efficiency of enlarging image segmentation datasets by using limited manual annotations, avoiding the exhaustive effort of full labeling. Recently, the Segment Anything Model (SAM)~\cite{kirillov2023segany} excels in real-time, high-quality interactive segmentation, responding to annotator prompts such as clicks~\cite{huang2023interformer}, bounding boxes~\cite{kirillov2023segany}, or coarse masks~\cite{KonstantinSofiiuk2021RevivingIT}. SAM's generalizability and efficiency in processing diverse prompts make it a versatile tool across a spectrum of segmentation-related tasks. This study focuses on click-based interactive segmentation building upon SAM~\cite{kirillov2023segany}.

SAM~\cite{kirillov2023segany} alongside the concurrent InterFormer~\cite{huang2023interformer} has pioneered a new interactive segmentation pipeline. This pipeline incorporates powerful Vision Transformers (ViTs)~\cite{AlexeyDosovitskiy2020AnII,YanghaoLiExploringPV,KaimingHe2021MaskedAA} as the image encoder to preprocess images, generating image embeddings that are applicable to all objects within the same image. During the interaction, these image embeddings and the prompts (\eg clicks) from annotators are fed into a lightweight decoder to produce segmentation results. This pipeline combines the power of large ViTs with the speed needed for on-the-spot interactive segmentation. Following such a pipeline, SAM even enables annotators to perform real-time, high-quality interactive segmentation on CPU-only devices, aiding in the significant expansion of image segmentation annotations~\cite{kirillov2023segany}.

However, SAM's pipeline has two limitations. First, the pipeline's image preprocessing disables the efficient implementation of the image-level zoom-in strategy~\cite{KonstantinSofiiuk2020fBRSRB} that dynamically refocuses the model on the target object during interaction.
Second, SAM's lightweight decoder struggles to sufficiently fuse the interactive information with the preprocessed image embeddings due to the need for real-time responses, thus weakening the interactive feedback's positive impact on segmentation quality. 
Consequently, SAM faces instability issues in challenging scenarios, such as camouflaged objects~\cite{fan2020camouflaged} almost blending into the background.
Figure~\ref{figure:motivation} clearly illustrates the instability of SAM's segmentation results, where an additional click following a sufficient number of previous ones (\eg, 9 clicks) can unexpectedly trigger substantial degradation in segmentation quality, exemplified by a drop in IoU from $88.59$ to $12.78$. Such instability significantly limits SAM's applicability in a broader range of image segmentation annotations.

Therefore, we propose FocSAM to address SAM's limitations. FocSAM's pipeline builds upon SAM and introduces an extra focus refiner. This refiner adjusts SAM's image embeddings for each object during the object's interaction, adding ignorable computations. The adjustment facilitates two major improvements. First, the refiner uses initial segmentation results to refocus the image embeddings on regions containing the target object, inspired by the image-level zoom-in~\cite{KonstantinSofiiuk2020fBRSRB}. Second, the refiner sufficiently fuses the embeddings with a few initial clicks that prove to have great impact on final segmentation results~\cite{ZhengLinFocusCutDI}, further enhancing the object-related embeddings.

To implement FocSAM's focus refiner with minimal computational overhead, we introduce Dynamic Window Multi-head Self-Attention (Dwin-MSA) and Pixel-wise Dynamic ReLU (P-DyReLU). Dwin-MSA partitions image embeddings into windows and perform efficient attention computations on a dynamic minimal subset of the windowed embeddings that intersect with previously predicted masks. Such a dynamic manner avoids redundant computations on irrelevant background areas. 
Dwin-MSA uses the shifting strategy~\cite{liu2022swin} to ensure long-distance interactions among embeddings, preserving dynamic efficiency.
P-DyReLU is employed as the non-linear activation in the Dwin-MSA to fuse the interactive information from a few initial clicks with the image embeddings. Specifically, P-DyReLU adopts DyReLU~\cite{chen2020dynamic} and utilizes SAM decoder's click-fused query embeddings to enhance the object-related image embeddings and suppress object-unrelated ones.

Experimentally, FocSAM demonstrates superior interactive segmentation performance over SAM with negligible additional computational costs. FocSAM matches the state-of-the-art SimpleClick~\cite{QinLiu2022SimpleClickII} in Number of Clicks (NoC) across datasets including DAVIS~\cite{FedericoPerazzi2016ABD}, SBD~\cite{BharathHariharan2011SemanticCF}, GrabCut~\cite{CarstenRother2004GrabCutIF}, Berkeley~\cite{KevinMcGuinness2010ACE}, MVTec~\cite{bergmann2019mvtec} and COD10K~\cite{fan2020camouflaged}, but FocSAM requires only about $5.6\%$ of the CPU inference time compared to SimpleClick. Moreover, as the number of objects per image surpasses $10$, FocSAM's time efficiency further improves, demanding roughly $1.2\%$ of the time required by SimpleClick for CPU inference.

We summarize our contributions as follows:
\begin{itemize}
    \item We introduce FocSAM to boost SAM's performance by dynamically enhancing the object-related image embeddings and deeply integrating interactive information into these embeddings.
    \item FocSAM is implemented by proposed Dwin-MSA and P-DyReLU with ignorable extra computational costs.
    \item FocSAM matches the state-of-the-art SimpleClick in NoC across datasets including DAVIS, SBD, GrabCut, Berkeley, MVTec and COD10K, requires just $5.6\%$ of SimpleClick's inference time on CPUs.
\end{itemize}
\section{Related Work}
\begin{figure*}
    \centering
    \includegraphics[width=0.96\linewidth]{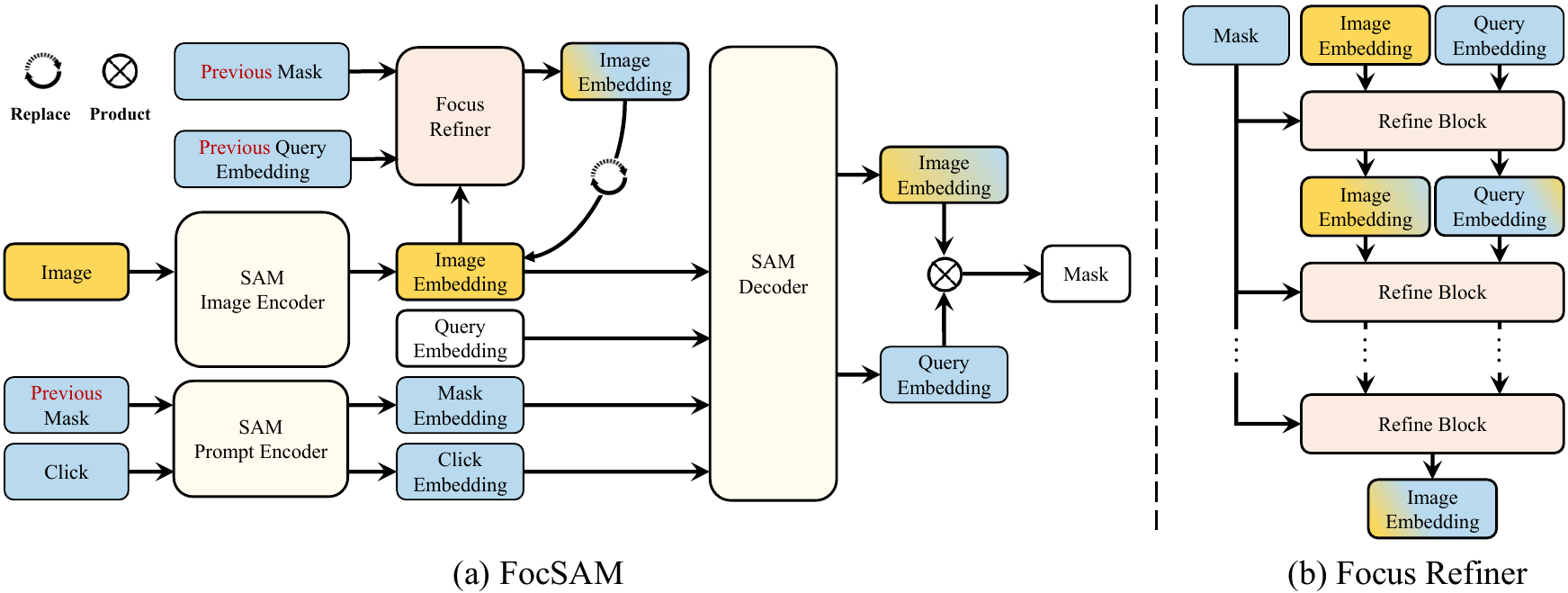}
    \vspace{-0.5em}
    \caption{
        Overview of proposed FocSAM building upon SAM. SAM comprises an image encoder, a prompt encoder and a decoder. The image encoder transforms images into image embeddings before interaction. In each interaction of an object, the prompt encoder converts the previous mask and annotator clicks into mask and click embeddings, respectively. These three embeddings and a learnable query embedding are fed into the decoder for segmentation. Upon SAM's pipeline, FocSAM introduces a focus refiner that is employed once per object during interaction (Figure (a)). In an early step of SAM's interaction, this refiner processes SAM's image embeddings, previous mask and click-fused query embedding through a stack of refine blocks (Figure (b)). Each block receives the image and query embeddings with the mask shared across all the blocks, and produces the image and query embeddings fed into the subsequent block. The final output is a refined image embedding, which replaces the original image embedding for subsequent interactions with the object.
    }
    \label{figure:focsam}
\end{figure*}

\label{sec:relatedwork}
\subsection{Interactive Segmentation}
The integration of deep networks into interactive segmentation~\cite{YuriBoykov2001InteractiveGC,LeoGrady2006RandomWF,VarunGulshan2010GeodesicSC,CarstenRother2004GrabCutIF} is initiated by DIOS~\cite{NingXu2016DeepIO}, leading to subsequent advancements in click-based methods like DEXTR~\cite{KevisKokitsiManinis2017DeepEC,ZhuwenLi2018InteractiveIS}, FCA-Net~\cite{ZhengLin2020InteractiveIS}, BRS~\cite{JangWonDong2019InteractiveIS}, and f-BRS~\cite{KonstantinSofiiuk2020fBRSRB}. The following methods~\cite{DavidAcuna2018EfficientIA,ShiyinZhang2020InteractiveOS,KonstantinSofiiuk2021RevivingIT,XiChen2022FocalClickTP,ZhengLinFocusCutDI,QinLiu2022PseudoClickII} focus on enhancing various aspects of interactive segmentation. SimpleClick~\cite{QinLiu2022SimpleClickII} is the first to introduce large Vision Transformers~\cite{AlexeyDosovitskiy2020AnII} into this field. InterFormer~\cite{huang2023interformer} follows with a novel pipeline to reduce model redundancy by reusing image features. SAM~\cite{kirillov2023segany} also adopts this pipeline and achieves robust zero-shot capabilities and diverse prompts, leading to various downstream applications~\cite{ma2023segment,mazurowski2023segment,lai2023lisa,wu2023medical,yu2023inpaint,wang2023review}. However, SAM is unable to employ the image-level zoom-in strategy~\cite{KonstantinSofiiuk2020fBRSRB} efficiently and integrate interactive information effectively, hindering its broader applications. We introduce FocSAM to address SAM's limitations.

\subsection{Efficient Attention}
Transformers~\cite{AshishVaswani2017AttentionIA} make remarkable strides in the field of computer vision~\cite{AlexeyDosovitskiy2020AnII,gu2022multi,khan2022transformers,RobinStrudel2021SegmenterTF,Xie_Wang_Yu_Anandkumar_Alvarez_Luo_2021,yuan2021hrformer,fang2023eva,wang2023internimage}. The high computational complexity of attention module leads to a range of research~\cite{liu2021swin,Zhuoran_Mingyuan_Haiyu_Shuai_Hongsheng_2021,Guo_Qiu_Liu_Shao_Xue_Zhang_2019,Xie_Wang_Yu_Anandkumar_Alvarez_Luo_2021}. One typical way is to limit the attention region of each token from full-attention to local/windowed attention~\cite{liu2021swin,Vaswani_Ramachandran_Srinivas_Parmar_Hechtman_Shlens_2021,YanghaoLiExploringPV,han2023flatten}. This strategy has garnered significant interest, as evidenced by various studies~\cite{Ho_Kalchbrenner_Weissenborn_Salimans_2019,Huang_Wang_Wei_Huang_Shi_Liu_Huang_2020,Chu_Tian_Wang_Zhang_Ren_Wei_Xia_Shen_2021,Wang_Yao_Chen_Cai_He_Liu_2021,Yang_Li_Zhang_Dai_Xiao_Yuan_Gao_Redmond_Cloud_Ai}. More recently, CSwin~\cite{Dong_Bao_Chen_Zhang_Yu_Yuan_Chen_Guo_2022} introduces Cross-Shaped Window Self-attention to compute concurrently in both orientations. Beyond Fixation~\cite{Ren_Li_Wang_Xiao_Chang} proposes DW-ViT to fuse multi-scale information. In this paper, we propose Dwin-MSA to perform dynamic window attention on object-related image embeddings.

\section{Method}
\begin{figure*}
    \centering
    \includegraphics[width=0.85\linewidth]{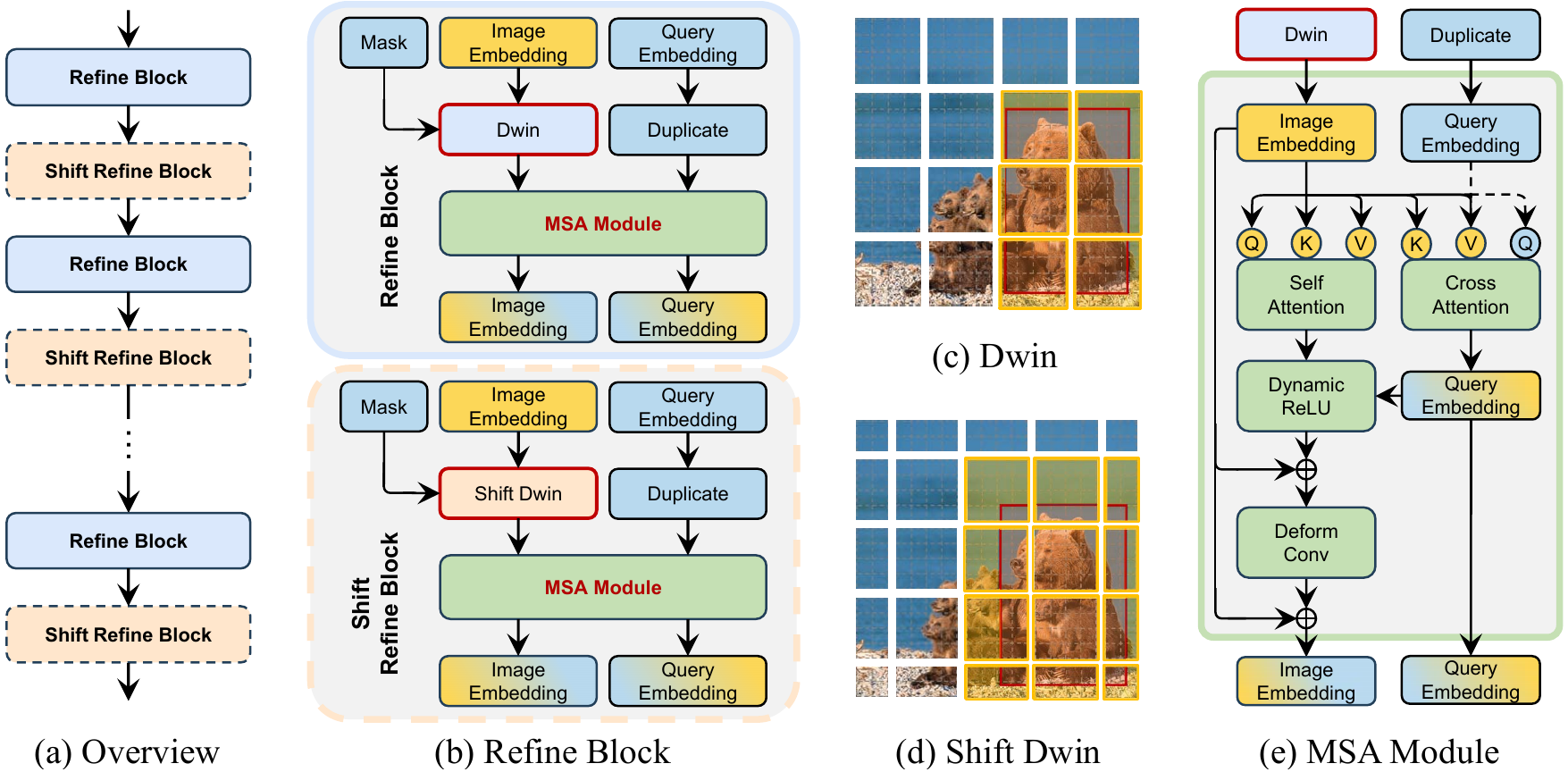}
    \caption{Overview of FocSAM's focus refiner. Figure (a) depicts the overall architecture of the focus refiner. Figure (b) details the refine block, showing the flow of image and query embeddings through the Dwin and MSA modules. Figures (c) and (d) highlight the window selection within the Dwin module and the shift strategy. Figure (e) provides a detailed view of the MSA module.}
    \label{figure:figure:focusrefiner}
    \vspace{-1.em}
\end{figure*}
We propose FocSAM with a redesigned SAM pipeline. In \ref{subsec:pipeline}, we present an overview of SAM's pipeline and the new pipeline. Then, we elaborate on the implementation of FocSAM's focus refiner in \ref{subsec:figure:focusrefiner} and \ref{subsec:pdyrelu}. Finally, the training loss is discussed in Section \ref{subsec:training}.

\subsection{Pipeline}
\label{subsec:pipeline}

\noindent\textbf{SAM's pipeline.} 
In Figure~\ref{figure:focsam}, SAM~\cite{kirillov2023segany} comprises an image encoder, a prompt encoder and a decoder. The image encoder preprocesses each image only once before the interaction, despite the varying number of objects within the image. Instead, both the prompt encoder and the decoder actively engage in every interaction, rapidly processing annotator clicks to predict segmentation results.

\noindent\textbf{Image encoder.} 
In SAM's preprocessing phase, images are resized and padded to $1024\times1024$ and fed into a ViT-based image encoder~\cite{AlexeyDosovitskiy2020AnII}. This encoder is structured in four stages of equal depth and utilizes window-based attention in each stage for efficient computation~\cite{YanghaoLiExploringPV}, with full attention applied at each stage's end. Following this, simple convolutional layers further reduces the dimensions to produce 256-dimensional embeddings $\boldsymbol{F} \in \mathop{\mathbb{R}}^{\frac{H}{16}\times \frac{W}{16} \times 256}$, corresponding to non-overlapping $16\times16$ image patches.

\noindent\textbf{Prompt encoder.} 
In SAM's interaction phase, the prompt encoder~\cite{kirillov2023segany} transforms annotator prompts into embeddings. These prompts include $N$ clicks at the $N$th interaction, each with $x, y$ coordinates and a label indicating positive or negative. A positive click in a false negative region signals the model to expand that region and a negative click in a false positive region suggests removal. Starting from the second interaction for each object, the prompt encoder also converts the previously predicted segmentation mask into mask embeddings. The transformed click embeddings $\boldsymbol{c} \in \mathop{\mathbb{R}}^{N\times 256}$ and mask embeddings $\boldsymbol{E} \in \mathop{\mathbb{R}}^{\frac{H}{16}\times \frac{W}{16} \times 256}$ will be fed into the SAM decoder, as depicted in Figure~\ref{figure:focsam}.

\noindent\textbf{Decoder.}
Following the prompt encoder, the decoder receives image embeddings $\boldsymbol{F}$, mask embeddings $\boldsymbol{E}$, click embeddings $\boldsymbol{c}$ and learnable query embeddings. The number of query embeddings corresponds to the expected output masks by the decoder. In our work, we use a single query embedding $\boldsymbol{q} \in \mathop{\mathbb{R}}^{1\times256}$. During decoding, the concatenated embeddings $[\boldsymbol{q}; \boldsymbol{c}] \in \mathop{\mathbb{R}}^{(N+1)\times 256}$ undergo cross-attention with the mask-fused image embedding $\boldsymbol{F} + \boldsymbol{E}$. They alternate roles of query and key/value in the cross attention without involving image-to-image attention. After two blocks of such cross-attention, the output includes the click-fused image embedding $\boldsymbol{F}_{c} \in \mathop{\mathbb{R}}^{\frac{H}{4}\times \frac{W}{4}\times 256}$ that has been upsampled by some convolutions and the click-fused query embedding $\boldsymbol{q}_{c} \in \mathop{\mathbb{R}}^{1\times 256}$, with the click embeddings discarded. Their dot product $\boldsymbol{F}_{c} \cdot \boldsymbol{q}_{c}^\top \in \mathop{\mathbb{R}}^{\frac{H}{4}\times \frac{W}{4} \times 1}$ generates logits for predicting the final mask $\boldsymbol{M}$. 

\noindent\textbf{FocSAM's pipeline.} 
Building upon SAM's pipeline, FocSAM's pipeline introduces the focus refiner. The refiner is employed once for each object. Specifically, at the  $K$th interaction of an object, the refiner receives the image embedding $\boldsymbol{F}$, the previously predicted mask $\boldsymbol{M}^{(K-1)}$ and the previous click-fused query embedding $\boldsymbol{q}_c^{(K - 1)}$. Then, the refiner produces a refined image embedding $\boldsymbol{F}_r^{(K)} \in \mathop{\mathbb{R}}^{\frac{H}{16}\times \frac{W}{16} \times 256}$ that has object-related embeddings. $\boldsymbol{F}_r^{(K)}$ replaces the original embedding $\boldsymbol{F}$ in all the subsequent interaction on this object. As illustrated by Figure~\ref{figure:focsam} (b), this focus refiner comprises a stack of refine blocks. These blocks refine the image and query embeddings iteratively, sharing the same previous mask. The image embedding from the final block serves as the refiner output. We detail these refine blocks in the following subsection. 

\subsection{Refine Block}
\label{subsec:figure:focusrefiner}

\noindent\textbf{Overview.}
In Figure~\ref{figure:figure:focusrefiner} (a), the plain refine block and the shift refine block alternately stack within the refiner, refining the image embedding $\boldsymbol{F}$ and click-fused query embedding $\boldsymbol{q}_c^{(K - 1)}$ with the shared mask $\boldsymbol{M}^{(K-1)}$. 
They share most modules, differing mainly in the Dwin and Shift Dwin (Figure~\ref{figure:figure:focusrefiner} (b)). 
Both the Dwin and Shift Dwin identify the bounding box around the object from the mask $\boldsymbol{M}^{(K-1)}$ (Figure~\ref{figure:figure:focusrefiner} (c)(d)) and refine the embeddings on the object. The refined embeddings and the correspondingly duplicated query embeddings are fed into the MSA module (Figure~\ref{figure:figure:focusrefiner} (e)). 
Then, we detail Dwin and Shift Dwin.

\noindent\textbf{Revisiting image-level zoom-in.}
Given an image $\mathcal{I}$ and a bounding box, the image-level zoom-in strategy~\cite{KonstantinSofiiuk2020fBRSRB} is
formulated as $\text{resize}(\mathcal{I}[y_1: y_2, x_1: x_2], (H, W))$, 
where corner coordinates $(x_1, y_1), (x_2, y_2)$ define the bounding box and $(H, W)$ is the model input size. Adapting this strategy to the embeddings typically involves RoIAlign~\cite{he2017mask} that crops and resizes embeddings using a linear sampling method. However, RoIAlign faces two main issues. First, RoIAlign assumes that embeddings can be linearly interpolated like images, which may not hold for SAM's image embeddings due to lack of the corresponding smoothness-aware training. 
Second, RoIAlign uniformly resizes all objects, ignoring size differences, which limits representation for larger objects and adds redundancy for smaller ones.

\noindent\textbf{Dynamic window.}
Instead of using RoIAlign, we introduce the Dynamic Window (Dwin) strategy. Given window size $S$, a batch of $B$ samples' image embeddings $\boldsymbol{F} \in \mathop{\mathbb{R}}^{B \times \frac{H}{16} \times \frac{W}{16} \times 256}$ can be windowed as $\bar{\boldsymbol{F}} \in \mathop{\mathbb{R}}^{L\times S \times S \times 256}$ with $L = BHW/(16S)^2$. Then, the windows intersecting the box are selected (Figure~\ref{figure:figure:focusrefiner} (c). For all objects within these images, we can simultaneously select all windows intersecting with their respective bounding boxes despite the objects' sizes. This leads to the selected embedding windows $\boldsymbol{F}_W \in \mathop{\mathbb{R}}^{M \times S \times S \times 256}$, with $M$ the number of windows interacting with the boxes. Each window performs independent computations like self-attention within the window, and updates its own embeddings with the computation results, freezing the unselected embedding windows.

\noindent\textbf{Long-range patch-to-patch attention.}
We further employ the shifting strategy~\cite{liu2021swin,liu2022swin} in the Shift Dwin (Figure~\ref{figure:figure:focusrefiner} (d)). Alternating the Dwin and Shift Dwin ensures sufficient information exchange between all the patches within the bounding box. Moreover, the boxes typically limit the spatial distance between embeddings within the same object, implying that a few blocks and small window sizes still allow sufficient information exchange.

\noindent\textbf{MSA module.}
The MSA (Figure~\ref{figure:figure:focusrefiner} (e)) processes $\boldsymbol{F}_W$'s each window \(\boldsymbol{f} \in \mathop{\mathbb{R}}^{S\times S\times 256}\) parallelly, with the duplicated query embedding \(\boldsymbol{q}_c = \text{copy}(\boldsymbol{q}_c^{K-1}) \in \mathop{\mathbb{R}}^{1\times 256}\). Let 
\vspace{-0.5em}
\begin{equation}
(Q, K, V)(\boldsymbol{x}, \boldsymbol{y}, \boldsymbol{z}) = \text{softmax}\left(\dfrac{\boldsymbol{x}W_QW_K^\top\boldsymbol{y}^\top}{\sqrt{d}}\right) \boldsymbol{z}W_V
\end{equation}
denote the conventional attention~\cite{AshishVaswani2017AttentionIA}. The MSA module is formulated as follows. First, \(\boldsymbol{q}_c\) is fused with \(\boldsymbol{f}\), \ie
\vspace{-0.5em}
\begin{equation}
    \boldsymbol{q}_f = (Q, K, V)(\boldsymbol{q}_c, \boldsymbol{f}, \boldsymbol{f}).
    \label{eq:qf}
\vspace{-0.5em}
\end{equation}
Then, \(\boldsymbol{f}\) undergoes self-attention, yielding
\vspace{-0.5em}
\begin{equation}
    \hat{\boldsymbol{\boldsymbol{f}}} = (Q, K, V)(\boldsymbol{f}, \boldsymbol{f}, \boldsymbol{f}).
    \label{eq:fhat}
\vspace{-0.5em}
\end{equation}
Next, \(\hat{\boldsymbol{f}}\) is activated by P-DyReLU as follows
\vspace{-0.5em}
\begin{equation}
    \hat{\boldsymbol{f}_q} = \text{PDyReLU}(\hat{\boldsymbol{f}}; \boldsymbol{q}_f).
    \label{eq:pdyrelu}
\vspace{-0.5em}
\end{equation}
Finally, this MSA module outputs both
\vspace{-0.5em}
\begin{equation}
    \boldsymbol{f}_q = \boldsymbol{f} + \text{DeformConv}(\boldsymbol{f} + \hat{\boldsymbol{f}_q})
\vspace{-0.5em}
\end{equation}
and \(\boldsymbol{q}_f\) as the next block's inputs. Additionally, the \(\boldsymbol{q}_f\) from each window is aggregated through an average summation. We detail P-DyReLU in the following subsection.

\subsection{Pixel-wise Dynamic ReLU}
\label{subsec:pdyrelu}

\noindent\textbf{Dynamic ReLU.}
DyReLU~\cite{chen2020dynamic} extends the conventional ReLU by introducing input-dependent activation parameters. For an input vector $\boldsymbol{x}$, the dynamic activation function $f(\boldsymbol{x}; {\boldsymbol{\theta}(\boldsymbol{x})})$ uses parameters $\boldsymbol{\theta}(\boldsymbol{x})$ that adapt based on $\boldsymbol{x}$. In details, the traditional ReLU function \(\boldsymbol{y}=\max\{\boldsymbol{x}, 0\}\) is generalized in DyReLU to a parametric piecewise linear function \(y_c=\max_k\{a_c^kx_c+b_c^k\}\) for each element $x_c$ of $\boldsymbol{x}$. DyReLU adapts coefficients \(a_c^k\) and \(b_c^k\) based on \(\boldsymbol{x}\):
\vspace{-0.5em}
\begin{equation}
y_c = f_{\boldsymbol{\theta}(\boldsymbol{x})}(x_c) = \max_{1 \leq k \leq K}\{a_c^k(\boldsymbol{x})x_c+b_c^k(\boldsymbol{x})\},
\vspace{-0.5em}
\end{equation}
where all the coefficients $\{a_c^k\}, \{b_c^k\}$ are outputs of the hyper function $\boldsymbol{\theta}(\boldsymbol{x})$. The plain ReLU is a special case of $K=2$ with $\boldsymbol{a}^1 = \boldsymbol{1}$ and $\boldsymbol{b}^1 = \boldsymbol{a}^2 = \boldsymbol{b}^2 = \boldsymbol{0}$.

\noindent\textbf{Pixel-wise DyReLU.}
Considering Equation~\ref{eq:pdyrelu}, we implement $\boldsymbol{\theta}(\boldsymbol{x})$ to fuse $\hat{\boldsymbol{f}} \in \mathop{\mathbb{R}}^{S\times S\times 256}$ from Equation~\ref{eq:qf} with $\boldsymbol{q}_f \in \mathop{\mathbb{R}}^{1 \times 256}$ from Equation~\ref{eq:fhat}. 
The implementation is inspired by the SAM decoders' use of a dot product between image and query embeddings to generate logits for mask prediction~\cite{kirillov2023segany}. 
This process effectively captures the unnormalized similarity between each image embedding and the query embedding in a pixel-wise manner. We adopt this similarity to enhance the object-related embeddings and suppress the unrelated ones, formulating $\boldsymbol{\theta}(\boldsymbol{x})$ as 
\vspace{-0.5em}
\begin{equation}
    \begin{split}
    &\boldsymbol{a}^0 = \boldsymbol{b}^0 = \text{Expand}(\hat{\boldsymbol{f}} \cdot \boldsymbol{q}_f^\top), \\
    &\boldsymbol{a}^1 = \boldsymbol{b}^1 = \text{Expand}(\text{AvgPool}(\hat{\boldsymbol{f}})), \\
    \end{split}
\end{equation}
where $\text{Expand}(\boldsymbol{x})$ replicate $\boldsymbol{x}$ to match the image embeddings $\hat{\boldsymbol{f}}$ and $\text{AvgPool}(\cdot)$ performs spatial average pooling. Thus, the coefficients $\boldsymbol{a}^0, \boldsymbol{a}^1, \boldsymbol{b}^0, \boldsymbol{b}^1$ share the same shape of $\hat{\boldsymbol{f}}$. Then, we apply channel-wise MLPs on these coefficients to transform their scales and bias, which yields
\vspace{-0.2em}
\begin{equation}
    \begin{split}
        &\bar{\boldsymbol{a}}^0 = \text{MLP}(\boldsymbol{a}^0; \boldsymbol{W}_a^0),
        \bar{\boldsymbol{b}}^0 = \text{MLP}(\boldsymbol{b}^0; \boldsymbol{W}_b^0),\\
        &\bar{\boldsymbol{a}}^1 = \text{MLP}(\boldsymbol{a}^1; \boldsymbol{W}_a^1),
        \bar{\boldsymbol{b}}^1 = \text{MLP}(\boldsymbol{b}^1; \boldsymbol{W}_b^1).
    \end{split}
\end{equation}
Finally, P-DyReLU in Equation~\ref{eq:pdyrelu} is implemented as
\vspace{-0.5em}
\begin{equation}
    \text{PDyReLU}(\hat{\boldsymbol{f}}; \boldsymbol{q}_f) = \max\{\bar{\boldsymbol{a}}^0 \odot \hat{\boldsymbol{f}} + \bar{\boldsymbol{b}}^0, \bar{\boldsymbol{a}}^1 \odot \hat{\boldsymbol{f}} + \bar{\boldsymbol{b}}^1\},
\vspace{-0.5em}
\end{equation}
where $\odot$ is an element-wise product.

\subsection{Training Loss}
\label{subsec:training}

Like previous methods~\cite{KonstantinSofiiuk2021RevivingIT,QinLiu2022SimpleClickII,huang2023interformer}, we adopt the normalized focal loss (NFL) proposed in RITM~\cite{KonstantinSofiiuk2021RevivingIT}. Additionally, we introduce the point loss (PTL) inspired by BRS~\cite{JangWonDong2019InteractiveIS} as the auxiliary loss, which is defined as follows
\vspace{-0.5em}
\begin{equation}
    \text{PTL}(\boldsymbol{M}, \{(x_i, y_i, z_i)\}) = \sum_{i} \left(\boldsymbol{M}(x_i, y_i) - z_i \right)^2,
\vspace{-0.5em}
\end{equation}
where $\{(x_i, y_i)\}$ is the coordinates of clicks leading to the predicted mask $\boldsymbol{M}$ and $z_i$ is the binary label indicating whether the click is positive.

\section{Experiments}
\begin{table*}[!ht]
    \centering
    \begin{tabular}{lcccccccc}
        \toprule
        Method & $\downarrow$SPC/s & GrabCut & Berkeley & SBD & DAVIS & MVTec & COD10K & Mean\\
        \toprule
        $\text{f-BRS-B-HR32~\cite{KonstantinSofiiuk2020fBRSRB} }_{\text{CVPR20}}$ & - & 1.69 & 2.44 & 7.26 & 6.50 & - & - & - \\ 
        $\text{RITM-HR18s~\cite{KonstantinSofiiuk2021RevivingIT} }_{\text{Preprint21}}$ & - & 1.68 & 2.60 & 6.48 & 5.98 & - & - & - \\
        $\text{RITM-HR32~\cite{KonstantinSofiiuk2021RevivingIT} }_{\text{Preprint21}}$ & - & 1.56 & 2.10 & 5.71 & 5.34 & - & - & - \\
        $\text{CDNet-R34~\cite{XiChen2021ConditionalDF} }_{\text{ICCV21}}$ & - & 1.52 & 2.06 & 7.04 & 5.56 & - & - & - \\
        $\text{EdgeFlow-HR18~\cite{YuyingHao2021EdgeFlowAP} }_{\text{ICCVW21}}$ & - & 1.72 & 2.40 & - & 5.77 & - & - & - \\
        $\text{PseudoClick-HR32~\cite{QinLiu2022PseudoClickII} }_{\text{ECCV22}}$ & - & 1.50 & 2.08 & 5.54 & 5.11 & - & - & - \\
        \midrule
        $\text{FocalClick-HR18s-S1~\cite{XiChen2022FocalClickTP} }_{\text{CVPR22}}$ & 0.03 & 1.82 & 2.89 & 7.29 & 6.56 & 13.99 & 13.39 & 7.66 \\
        $\text{FocalClick-HR18s-S2~\cite{XiChen2022FocalClickTP} }_{\text{CVPR22}}$ & 0.07 & 1.62 & 2.66 & 6.79 & 5.25 & 13.29 & 12.00 & 6.93 \\
        $\text{FocalClick-HR32-S2~\cite{XiChen2022FocalClickTP} }_{\text{CVPR22}}$ & 0.14 & 1.80 & 2.36 & 6.51 & 5.39 & 12.40 & 11.59 & 6.67 \\
        $\text{FocalClick-SegFB0-S1~\cite{XiChen2022FocalClickTP} }_{\text{CVPR22}}$ & 0.01 & 1.86 & 3.29 & 7.60 & 7.42 & 13.99 & 14.01 & 8.03 \\
        $\text{FocalClick-SegFB0-S2~\cite{XiChen2022FocalClickTP} }_{\text{CVPR22}}$ & 0.02 & 1.66 & 2.27 & 6.86 & 5.49 & 12.31 & 11.77 & 6.73 \\
        $\text{FocalClick-SegFB3-S2~\cite{XiChen2022FocalClickTP} }_{\text{CVPR22}}$ & 0.10 & 1.50 & 1.92 & 5.59 & 4.90 & 11.20 & 10.54 & 5.94 \\
        \midrule
        $\text{InterFormer-Light~\cite{huang2023interformer} }_{\text{ICCV23}}$ & 0.13 (0.10)$^\dagger$ & 1.50 & 3.14 & 6.34 & 6.19 & 12.03 & 11.27 & 6.75 \\
        $\text{InterFormer-Tiny~\cite{huang2023interformer} }_{\text{ICCV23}}$ & 0.23 (0.14)$^\dagger$ & 1.36 & 2.53 & 5.51 & 5.21 & 10.84 & 9.42 & 5.81 \\
        \midrule
        $\text{SimpleClick-ViT-B~\cite{QinLiu2022SimpleClickII} }_{\text{ICCV23}}$ & 1.26 & 1.48 & 1.97 & 5.62 & 5.06 & 11.15 & 9.93 & 5.87 \\
        $\text{SimpleClick-ViT-L~\cite{QinLiu2022SimpleClickII} }_{\text{ICCV23}}$ & 3.12 & 1.40 & 1.89 & 4.89 & 4.81 & 10.65 & 9.07 & 5.45 \\
        $\text{SimpleClick-ViT-H~\cite{QinLiu2022SimpleClickII} }_{\text{ICCV23}}$ & 6.99 & 1.50 & 1.75 & 4.70 & 4.78 & \textbf{10.56} & 9.13 & 5.40 \\
        \midrule
        $^{\ddagger}\text{SAM-ViT-H~\cite{kirillov2023segany} }_{\text{ICCV23} }$ & 0.35 (0.02)$^\dagger$ & 1.88 & 2.09 & 7.62 & 5.19 & 13.97 & 10.36 & 6.85 \\
        $\text{FocSAM-ViT-H (Ours)}$ & 0.39 (0.02)$^\dagger$ & \textbf{1.32} & \textbf{1.47} & \textbf{4.69} & \textbf{4.77} & 11.14 & \textbf{8.91} & \textbf{5.38} \\
        \bottomrule
    \end{tabular}
    \caption{Comparison of NoC@90 with previous methods. We report results on GrabCut~\cite{CarstenRother2004GrabCutIF}, Berkeley~\cite{KevinMcGuinness2010ACE}, SBD~\cite{BharathHariharan2011SemanticCF}, DAVIS~\cite{FedericoPerazzi2016ABD}, MVTec~\cite{bergmann2019mvtec} and COD10K~\cite{fan2020camouflaged}. The best results are highlighted in bold. 
    \(\dagger\) signifies that the SPC metric incorporates both decoder inference time and encoder inference time averaged over 20 clicks. For our FocSAM, the SPC additionally includes the proposed refiner's inference time averaged over 20 clicks. The decoder-only SPC is separately noted in parentheses, indicating the actual interaction time.
    \(\ddagger\) denotes methods that have not followed the conventional COCO~\cite{TsungYiLin2014MicrosoftCC}+LVIS~\cite{AgrimGupta2019LVISAD} training for interactive segmentation. Our FocSAM achieves state-of-the-art NoC@90 performance, while the SPC on CPUs is only about $5.6\%$ of the previous SOTA SimpleClick-ViT-H~\cite{QinLiu2022SimpleClickII}.
    }
    \label{tab:noc90s}
\end{table*}

In Section~\ref{sec:experimentalsetting}, we detail the experimental setup. Section~\ref{sec:mainresult} discusses the main results, comparing FocSAM's performance with previous methods across various datasets. In Section~\ref{sec:stabilityanalysis}, we statistically evaluate the stability of FocSAM in interactive segmentation, compared to SAM. The impact of FocSAM's modules is explored in Section~\ref{sec:ablationstudy}. Finally, Section~\ref{sec:qualitativeresult} presents qualitative results.

\subsection{Experimental Setting}
\label{sec:experimentalsetting}

\noindent\textbf{Datasets.} 
Following the previous methods~\cite{XiChen2022FocalClickTP,QinLiu2022PseudoClickII,QinLiu2022SimpleClickII,huang2023interformer}, we train our models on COCO~\cite{TsungYiLin2014MicrosoftCC} and LVIS~\cite{AgrimGupta2019LVISAD}, and then evaluate all the methods' zero-shot interactive segmentation capabilities on various other datasets including GrabCut~\cite{CarstenRother2004GrabCutIF}, Berkeley~\cite{KevinMcGuinness2010ACE}, SBD~\cite{BharathHariharan2011SemanticCF} and DAVIS~\cite{FedericoPerazzi2016ABD}. Our evaluation also extends to more challenging datasets including MVTec~\cite{bergmann2019mvtec} and COD10K~\cite{fan2020camouflaged}. 
Please refer to the supplementary materials for more details on the datasets.

\noindent\textbf{Implementation details.}
We utilize the pre-trained ViT-Huge from SAM~\cite{kirillov2023segany} as the backbone with the prompt encoder and decoder. For the proposed focus refiner, we configure a total of $12$ blocks, comprising $6$ plain refine blocks and $6$ shift refine blocks. The embedding dimensions of both Dwin-MSA and P-DyReLU are set to align with the $256$-dimensional SAM image embeddings. The window size for Dwin-MSA is set to $16$. The refine step $K$ is set to $2$, \ie, the focus refiner activates after the second click. Further details are available in the supplementary materials.

\begin{figure*}
    \centering
    \includegraphics[width=1.0\linewidth]{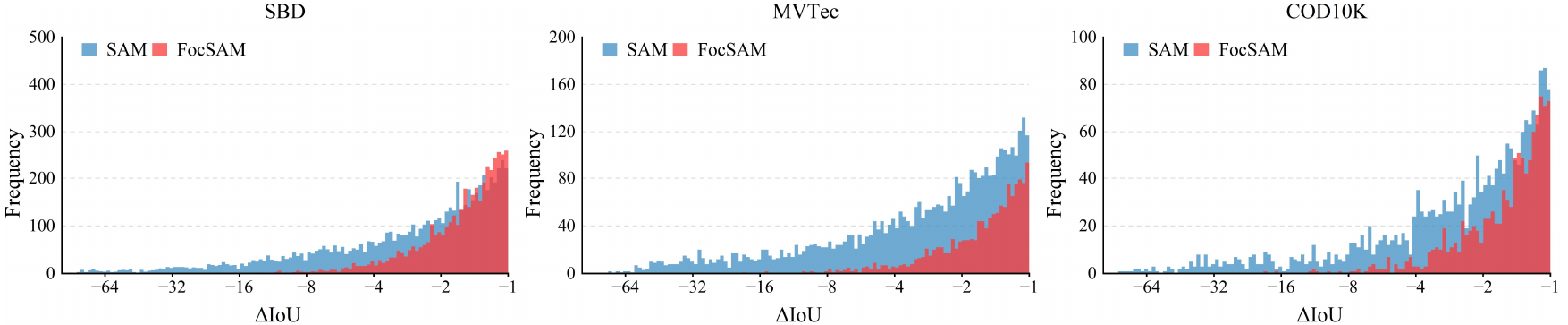}
    \caption{Stability analysis of interactive segmentation. We report results on SBD~\cite{BharathHariharan2011SemanticCF}, MVTec~\cite{bergmann2019mvtec} and COD10K~\cite{fan2020camouflaged}, and show $\Delta$IoU for consecutive clicks, filtering out $\Delta$IoU greater than $-1\%$. The results highlight FocSAM's superior stability over SAM, evidenced by fewer significant declines in segmentation quality with additional clicks.
    }
    \label{figure:stats}
\end{figure*}

\begin{table*}[ht]
    \small
    \centering
    \begin{tabular}{c c c c c c c c}
        \toprule
        \multirow{2}{*}{Dwin-MSA} & \multirow{2}{*}{P-DyReLU} & \multicolumn{2}{c}{SBD} & \multicolumn{2}{c}{MVTec} & \multicolumn{2}{c}{COD10K} \\
         &  & 20NoC@90 & 100NoC@95 & 20NoC@90 & 100NoC@95 & 20NoC@90 & 100NoC@95 \\ 
        \midrule
        \ding{55} & \ding{55} & 7.62 & 63.40 & 13.97 & 81.90 & 10.36 & 76.73 \\
        \ding{51} & \ding{55}  & 4.75 & 34.39 & 11.29 & 64.15 & 9.26 & 64.32 \\
        \ding{55} & \ding{51} & 4.76 & 34.52 & 11.48 & 65.04 & 9.33 & 64.41 \\
        \ding{51} & \ding{51} & 4.69 & 32.96 & 11.14 & 62.82 & 8.91 & 62.61 \\
        \bottomrule
    \end{tabular}
    \caption{Ablation study on Dwin-MSA and P-DyReLU. We measure NoC@90 with up to 20 clicks (20NoC@90) and NoC@95 with up to 100 clicks (100NoC@95). Our findings reveal: 1) the metric under 100 clicks emphasizes the influence of challenging samples; 2) Dwin-MSA and P-DyReLU individually yield similar results; 3) combining Dwin-MSA with P-DyReLU enhances the performance, especially evident under 100 clicks, which reduces the negative impact of challenging samples.}
    \label{tab:ablation}
\end{table*}

\noindent\textbf{Training strategy.}
In training FocSAM, we adopt InterFormer's click simulation strategy~\cite{huang2023interformer} for interactive simulation before loss computation. SAM's image encoder and prompt encoder are frozen during training. Moreover, we use the image encoder to pre-extract and store the COCO-LVIS image embeddings to reduce computational costs. We resize and pad the images to match SAM's input size of $1024\times1024$. We employ a two-stage training strategy involving firstly fine-tuning the SAM decoder for $320$k iterations at a batch size of $16$ and then training FocSAM with the frozen decoder for $160$k iterations in the same settings. This strategy addresses the training instability caused by the refiner's loss dependency on the decoder. Training and evaluations are performed on a server with $4$ NVIDIA RTX 3090 GPUs and dual Intel Xeon Silver CPUs. More details are provided in the supplementary materials.

\noindent\textbf{Evaluation.}
In the evaluation, following SAM~\cite{kirillov2023segany}, images are resized and padded to $1024$, and the segmentation results from the decoder are then adjusted back to their original size for IoU calculations. For click simulation in testing, we place clicks at the centers of erroneously predicted regions, in line with previous methods~\cite{XiChen2022FocalClickTP,QinLiu2022SimpleClickII,huang2023interformer}. The binary label of each click is determined by the maximum distance to the boundaries of false negative and false positive regions. FocSAM is evaluated in both inference speed and segmentation performance. Speed is quantified as Seconds Per Click (SPC) on CPUs, indicating the average inference time per click. For segmentation performance, we use the Number of Clicks (NoC) metric that is the average minimum clicks required to reach a specified IoU. We mainly focus on NoC@90 under 20 clicks, \ie, the average clicks needed to achieve $90\%$ IoU. In cases where more than 20 clicks are needed, the count is capped at 20 for evaluation consistency with previous methods~\cite{XiChen2022FocalClickTP,QinLiu2022SimpleClickII,huang2023interformer}. Additional NoC metrics are employed in the ablation study.

\begin{figure*}
    \centering
    \includegraphics[width=0.95\linewidth]{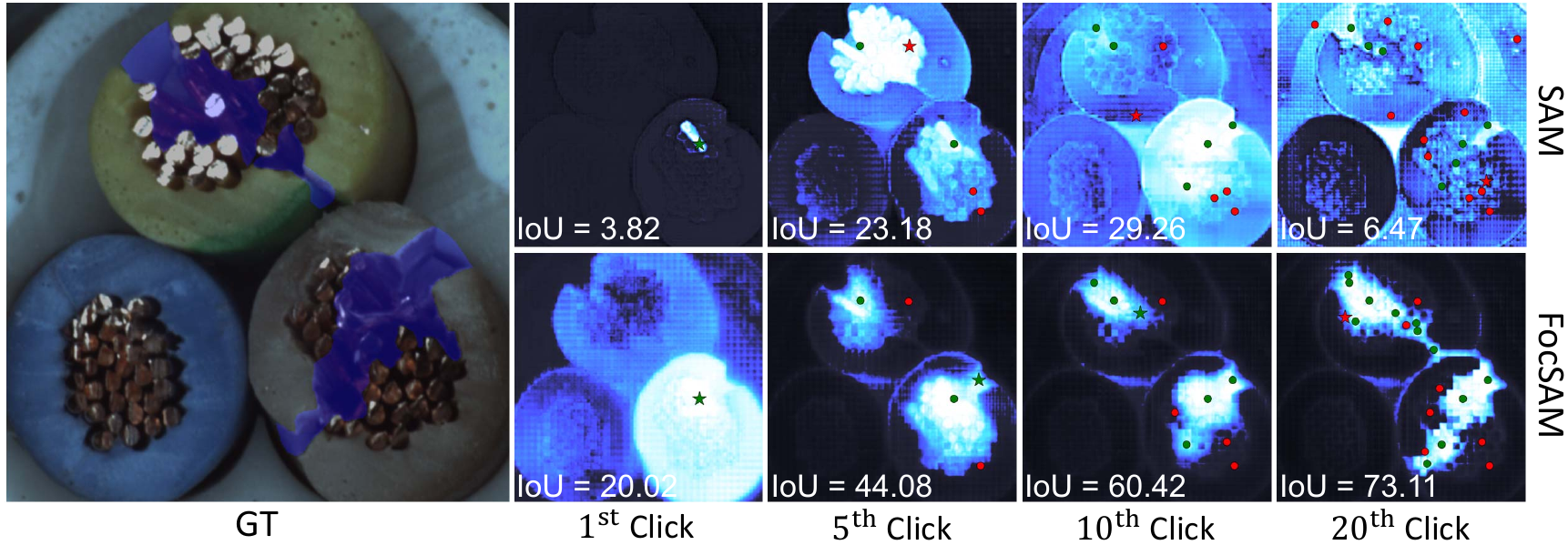}
    \caption{Qualitative analysis on a challenge Example. The first image from the left displays the challenge example with the image and GT (blue masks). The top and bottom rows on the right respectively show the segmentation results of SAM and FocSAM at the $1^\text{st}$, $5^\text{th}$, $10^\text{th}$, and $20^\text{th}$ clicks. Clicks are indicated with green (positive) and red (negative) circles.}
    \label{figure:qualitative}
\end{figure*}

\subsection{Main Results}
\label{sec:mainresult}

Table~\ref{tab:noc90s} showcases FocSAM's main results, benchmarked against previous methods. 
Indeed, SAM has been benchmarked against the mainstream methods~\cite{KonstantinSofiiuk2021RevivingIT, QinLiu2022SimpleClickII, XiChen2022FocalClickTP} in its experiments~\cite{kirillov2023segany} despite SAM's pretraining on SA-1B~\cite{kirillov2023segany} instead of COCO+LVIS used for these methods. 
The SA-1B and COCO+LVIS are both designed for general scenarios and often overlap in scope, facilitating valid comparisons between SAM and these methods. 
Due to SA-1B's inclusion of numerous SAM-generated masks, FocSAM maintains training on COCO+LVIS to mitigate bias inherent in SAM.
As reported, FocSAM achieves state-of-the-art performance in five out of the six evaluation datasets, particularly in the largest SBD ($6671$ samples) and the second-largest COD10K datasets ($2026$ samples). Despite a slight underperformance in the MVTec dataset, FocSAM still maintains the best average NoC across all datasets, closely match the previous state-of-the-art SimpleClick-ViT-H~\cite{QinLiu2022SimpleClickII}. However, the standout aspect of FocSAM is its time efficiency, evidenced by an SPC of $0.39$, far quicker than SimpleClick-ViT-H's $6.99$ SPC. This is attributed to FocSAM's use of SAM's pipeline, which pre-extracts image embeddings for efficient interaction, unlike SimpleClick's full model inference at each interaction. On the other hand, although SAM shows slightly less inference time, its segmentation performance is lower compared to the early methods like FocalClick~\cite{XiChen2022FocalClickTP}. FocSAM enhances SAM's performance to top-tier levels in interactive segmentation while adding only about $10\%$ computational costs. The following subsection will further validate whether FocSAM genuinely resolves the instability issues in SAM.

\subsection{Stability Analysis}
\label{sec:stabilityanalysis}

\noindent\textbf{Experimental Settings.}  
To evaluate the stability, we conduct statistical analyses on the three large-scale datasets, \ie SBD, MVTec and COD10K. Similar to the evaluation on NoC metrics, each click is placed at the center of the erroneously predicted regions. The number of simulated clicks per sample is increased from $20$ to $100$. For each sample, from the second interaction click onwards, we calculate the $\Delta$IoU, which is the difference in IoU between consecutive clicks, and filter out $\Delta$IoU greater than $-1\%$. This ensures that only significant deteriorations in segmentation quality are considered. The remaining $\Delta$IoUs are then visualized.

\noindent\textbf{Results.}  
As illustrated in Figure~\ref{figure:stats}, FocSAM exhibits considerably better stability across all datasets compared to SAM. The $\Delta$IoU distribution for FocSAM shows a rightward shift, indicating fewer samples of deteriorating segmentation with subsequent clicks. 
Although SAM occasionally achieves favorable outcomes, its inherent instability often necessitates additional annotator interactions for correcting errors. Therefore, FocSAM represents a stability advance over SAM in terms of real-world interactive efficiency, as evidenced by the stability analysis.

\subsection{Ablation Study}
\label{sec:ablationstudy}
\noindent\textbf{Experimental Settings.}  
In the ablation study, we evaluate the individual impact of Dwin-MSA and P-DyReLU on FocSAM's performance. Due to the interdependence of these modules, we slightly modify the modules. For Dwin-MSA only, we remove all P-DyReLU modules, replacing P-DyReLU's activations in Dwin-MSA with standard ReLU. For P-DyReLU only, we remove the dynamic windows to retain all image embeddings, and remove Dwin's attention computations. We evaluate these variants on the three largest datasets including SBD, MVTec, and COD10K, using NoC@90 within 20 clicks, and extend to NoC@95 within 100 clicks for deeper analysis. This NoC@95 metric quantifies the individual contributions of each module, especially on more challenging samples. All ablation models are trained with the same protocol of the main experiments.

\noindent\textbf{Results.}  
Table~\ref{tab:ablation} shows that Dwin-MSA and P-DyReLU individually contribute similarly to FocSAM's performance, indicating that they provide comparable interactive information. Dwin-MSA primarily focuses on initially predicted masks for locating main object areas, similar to bounding box prompts in SAM, whereas P-DyReLU leverages initial clicks for primary object outlining. Their interactive information is complementary. Consequently, their combination leads to enhanced overall performance, particularly noticeable in NoC@95 under 100 clicks. This metric underscores the increased click requirement to achieve $95\%$ IoU for challenging samples. The integration of Dwin-MSA and P-DyReLU further stabilizes FocSAM's performance on challenging samples. More ablation studies are provided in the supplementary materials.

\subsection{Qualitative Results}
\label{sec:qualitativeresult}
In Figure~\ref{figure:qualitative}, we present a qualitative comparison of FocSAM and SAM using a challenging example and visualize the segmentation results at four different clicks. This visualization clearly demonstrates FocSAM's enhanced stability over SAM. Our qualitative analysis confirms that FocSAM maintains consistent performance, providing superior segmentation quality compared to SAM under such a challenging example. Additional qualitative results are available in the supplementary materials.

\section{Conclusion}
SAM provides an efficient real-time pipeline for interactive segmentation, significantly advancing this field. However, SAM's real-world application stability is compromised, particularly in challenging scenarios. This instability largely stems from SAM's pipeline, which lacks the capability to effectively focus on the target object. Our proposed FocSAM tackles these stability issues by redesigning the pipeline to dynamically refocus SAM's image embeddings onto the target object. This adaptation enables FocSAM to stabilize the interactive segmentation process of SAM, even in challenging scenarios. As a result, FocSAM not only matches the state-of-the-art in segmentation quality but also achieves this with considerably lower computational demands on CPUs. These advancements highlight FocSAM's potential for broader real-world application.

\section*{Acknowledgments}
This work was supported by National Science and Technology Major Project (No. 2022ZD0118202), the National Science Fund for Distinguished Young Scholars (No.62025603), the National Natural Science Foundation of China (No. U21B2037, No. U22B2051, No. 62176222, No. 62176223, No. 62176226, No. 62072386, No. 62072387, No. 62072389, No. 62002305 and No. 62272401), and the Natural Science Foundation of Fujian Province of China (No.2021J01002,  No.2022J06001).

{
    \small
    \bibliographystyle{ieeenat_fullname}
    \bibliography{main}

\begin{thebibliography}{60}
\providecommand{\natexlab}[1]{#1}
\providecommand{\url}[1]{\texttt{#1}}
\expandafter\ifx\csname urlstyle\endcsname\relax
  \providecommand{\doi}[1]{doi: #1}\else
  \providecommand{\doi}{doi: \begingroup \urlstyle{rm}\Url}\fi

\bibitem[Acuna et~al.(2018)Acuna, Ling, Kar, and
  Fidler]{DavidAcuna2018EfficientIA}
David Acuna, Huan Ling, Amlan Kar, and Sanja Fidler.
\newblock Efficient interactive annotation of segmentation datasets with
  polygon-rnn++.
\newblock In \emph{2018 {IEEE} Conference on Computer Vision and Pattern
  Recognition, {CVPR} 2018, Salt Lake City, UT, USA, June 18-22, 2018}, pages
  859--868. {IEEE} Computer Society, 2018.

\bibitem[Bergmann et~al.(2019)Bergmann, Fauser, Sattlegger, and
  Steger]{bergmann2019mvtec}
Paul Bergmann, Michael Fauser, David Sattlegger, and Carsten Steger.
\newblock Mvtec ad--a comprehensive real-world dataset for unsupervised anomaly
  detection.
\newblock In \emph{Proceedings of the IEEE/CVF conference on computer vision
  and pattern recognition}, pages 9592--9600, 2019.

\bibitem[Boykov and Jolly(2001)]{YuriBoykov2001InteractiveGC}
Yuri~Y Boykov and M-P Jolly.
\newblock Interactive graph cuts for optimal boundary \& region segmentation of
  objects in nd images.
\newblock 1:\penalty0 105--112, 2001.

\bibitem[Chen et~al.(2021)Chen, Zhao, Yu, Zhang, and
  Duan]{XiChen2021ConditionalDF}
Xi Chen, Zhiyan Zhao, Feiwu Yu, Yilei Zhang, and Manni Duan.
\newblock Conditional diffusion for interactive segmentation.
\newblock pages 7345--7354, 2021.

\bibitem[Chen et~al.(2022)Chen, Zhao, Zhang, Duan, Qi, and
  Zhao]{XiChen2022FocalClickTP}
Xi Chen, Zhiyan Zhao, Yilei Zhang, Manni Duan, Donglian Qi, and Hengshuang
  Zhao.
\newblock Focalclick: towards practical interactive image segmentation.
\newblock pages 1300--1309, 2022.

\bibitem[Chen et~al.(2020)Chen, Dai, Liu, Chen, Yuan, and Liu]{chen2020dynamic}
Yinpeng Chen, Xiyang Dai, Mengchen Liu, Dongdong Chen, Lu Yuan, and Zicheng
  Liu.
\newblock Dynamic relu.
\newblock In \emph{European Conference on Computer Vision}, pages 351--367.
  Springer, 2020.

\bibitem[Chu et~al.(2021)Chu, Tian, Wang, Zhang, Ren, Wei, Xia, and
  Shen]{Chu_Tian_Wang_Zhang_Ren_Wei_Xia_Shen_2021}
Xiangxiang Chu, Zhi Tian, Yuqing Wang, Bo Zhang, Haibing Ren, Xiaolin Wei,
  Huaxia Xia, and Chunhua Shen.
\newblock Twins: Revisiting the design of spatial attention in vision
  transformers.
\newblock \emph{Neural Information Processing Systems,Neural Information
  Processing Systems}, 2021.

\bibitem[Dong et~al.(2022)Dong, Bao, Chen, Zhang, Yu, Yuan, Chen, and
  Guo]{Dong_Bao_Chen_Zhang_Yu_Yuan_Chen_Guo_2022}
Xiaoyi Dong, Jianmin Bao, Dongdong Chen, Weiming Zhang, Nenghai Yu, Lu Yuan,
  Dong Chen, and Baining Guo.
\newblock Cswin transformer: A general vision transformer backbone with
  cross-shaped windows.
\newblock In \emph{2022 IEEE/CVF Conference on Computer Vision and Pattern
  Recognition (CVPR)}, 2022.

\bibitem[Dosovitskiy et~al.(2021)Dosovitskiy, Beyer, Kolesnikov, Weissenborn,
  Zhai, Unterthiner, Dehghani, Minderer, Heigold, Gelly, Uszkoreit, and
  Houlsby]{AlexeyDosovitskiy2020AnII}
Alexey Dosovitskiy, Lucas Beyer, Alexander Kolesnikov, Dirk Weissenborn,
  Xiaohua Zhai, Thomas Unterthiner, Mostafa Dehghani, Matthias Minderer, Georg
  Heigold, Sylvain Gelly, Jakob Uszkoreit, and Neil Houlsby.
\newblock An image is worth 16x16 words: Transformers for image recognition at
  scale.
\newblock In \emph{9th International Conference on Learning Representations,
  {ICLR} 2021, Virtual Event, Austria, May 3-7, 2021}. OpenReview.net, 2021.

\bibitem[Fan et~al.(2020)Fan, Ji, Sun, Cheng, Shen, and
  Shao]{fan2020camouflaged}
Deng-Ping Fan, Ge-Peng Ji, Guolei Sun, Ming-Ming Cheng, Jianbing Shen, and Ling
  Shao.
\newblock Camouflaged object detection.
\newblock In \emph{Proceedings of the IEEE/CVF conference on computer vision
  and pattern recognition}, pages 2777--2787, 2020.

\bibitem[Fang et~al.(2023)Fang, Wang, Xie, Sun, Wu, Wang, Huang, Wang, and
  Cao]{fang2023eva}
Yuxin Fang, Wen Wang, Binhui Xie, Quan Sun, Ledell Wu, Xinggang Wang, Tiejun
  Huang, Xinlong Wang, and Yue Cao.
\newblock Eva: Exploring the limits of masked visual representation learning at
  scale.
\newblock In \emph{Proceedings of the IEEE/CVF Conference on Computer Vision
  and Pattern Recognition}, pages 19358--19369, 2023.

\bibitem[Grady(2006)]{LeoGrady2006RandomWF}
Leo Grady.
\newblock Random walks for image segmentation.
\newblock \emph{IEEE Transactions on Pattern Analysis and Machine
  Intelligence}, 2006.

\bibitem[Gu et~al.(2022)Gu, Kwon, Wang, Ye, Li, Chen, Lai, Chandra, and
  Pan]{gu2022multi}
Jiaqi Gu, Hyoukjun Kwon, Dilin Wang, Wei Ye, Meng Li, Yu-Hsin Chen, Liangzhen
  Lai, Vikas Chandra, and David~Z Pan.
\newblock Multi-scale high-resolution vision transformer for semantic
  segmentation.
\newblock In \emph{Proceedings of the IEEE/CVF Conference on Computer Vision
  and Pattern Recognition}, pages 12094--12103, 2022.

\bibitem[Gulshan et~al.(2010)Gulshan, Rother, Criminisi, Blake, and
  Zisserman]{VarunGulshan2010GeodesicSC}
Varun Gulshan, Carsten Rother, Antonio Criminisi, Andrew Blake, and Andrew
  Zisserman.
\newblock Geodesic star convexity for interactive image segmentation.
\newblock In \emph{The Twenty-Third {IEEE} Conference on Computer Vision and
  Pattern Recognition, {CVPR} 2010, San Francisco, CA, USA, 13-18 June 2010},
  pages 3129--3136. {IEEE} Computer Society, 2010.

\bibitem[Guo et~al.(2019)Guo, Qiu, Liu, Shao, Xue, and
  Zhang]{Guo_Qiu_Liu_Shao_Xue_Zhang_2019}
Qipeng Guo, Xipeng Qiu, Pengfei Liu, Yunfan Shao, Xiangyang Xue, and Zheng
  Zhang.
\newblock Star-transformer.
\newblock In \emph{Proceedings of the 2019 Conference of the North}, 2019.

\bibitem[Gupta et~al.(2019)Gupta, Doll{\'{a}}r, and
  Girshick]{AgrimGupta2019LVISAD}
Agrim Gupta, Piotr Doll{\'{a}}r, and Ross~B. Girshick.
\newblock {LVIS:} {A} dataset for large vocabulary instance segmentation.
\newblock In \emph{{IEEE} Conference on Computer Vision and Pattern
  Recognition, {CVPR} 2019, Long Beach, CA, USA, June 16-20, 2019}, pages
  5356--5364. Computer Vision Foundation / {IEEE}, 2019.

\bibitem[Han et~al.(2023)Han, Pan, Han, Song, and Huang]{han2023flatten}
Dongchen Han, Xuran Pan, Yizeng Han, Shiji Song, and Gao Huang.
\newblock Flatten transformer: Vision transformer using focused linear
  attention.
\newblock In \emph{Proceedings of the IEEE/CVF International Conference on
  Computer Vision}, pages 5961--5971, 2023.

\bibitem[Hao et~al.(2021)Hao, Liu, Wu, Han, Chen, Chen, Chu, Tang, Yu, Chen,
  et~al.]{YuyingHao2021EdgeFlowAP}
Yuying Hao, Yi Liu, Zewu Wu, Lin Han, Yizhou Chen, Guowei Chen, Lutao Chu,
  Shiyu Tang, Zhiliang Yu, Zeyu Chen, et~al.
\newblock Edgeflow: Achieving practical interactive segmentation with
  edge-guided flow.
\newblock pages 1551--1560, 2021.

\bibitem[Hariharan et~al.(2011)Hariharan, Arbelaez, Bourdev, Maji, and
  Malik]{BharathHariharan2011SemanticCF}
Bharath Hariharan, Pablo Arbelaez, Lubomir~D. Bourdev, Subhransu Maji, and
  Jitendra Malik.
\newblock Semantic contours from inverse detectors.
\newblock In \emph{{IEEE} International Conference on Computer Vision, {ICCV}
  2011, Barcelona, Spain, November 6-13, 2011}, pages 991--998. {IEEE} Computer
  Society, 2011.

\bibitem[He et~al.(2017)He, Gkioxari, Doll{\'a}r, and Girshick]{he2017mask}
Kaiming He, Georgia Gkioxari, Piotr Doll{\'a}r, and Ross Girshick.
\newblock Mask r-cnn.
\newblock In \emph{Proceedings of the IEEE international conference on computer
  vision}, pages 2961--2969, 2017.

\bibitem[He et~al.(2021)He, Chen, Xie, Li, Doll{\'a}r, and
  Girshick]{KaimingHe2021MaskedAA}
Kaiming He, Xinlei Chen, Saining Xie, Yanghao Li, Piotr Doll{\'a}r, and Ross
  Girshick.
\newblock Masked autoencoders are scalable vision learners.
\newblock \emph{arXiv: Computer Vision and Pattern Recognition}, 2021.

\bibitem[Ho et~al.(2019)Ho, Kalchbrenner, Weissenborn, and
  Salimans]{Ho_Kalchbrenner_Weissenborn_Salimans_2019}
Jonathan Ho, Nal Kalchbrenner, Dirk Weissenborn, and Tim Salimans.
\newblock Axial attention in multidimensional transformers.
\newblock \emph{Cornell University - arXiv,Cornell University - arXiv}, 2019.

\bibitem[Huang et~al.(2023)Huang, Yang, Sun, Zhang, Cao, Jiang, and
  Ji]{huang2023interformer}
You Huang, Hao Yang, Ke Sun, Shengchuan Zhang, Liujuan Cao, Guannan Jiang, and
  Rongrong Ji.
\newblock Interformer: Real-time interactive image segmentation.
\newblock In \emph{Proceedings of the IEEE/CVF International Conference on
  Computer Vision}, pages 22301--22311, 2023.

\bibitem[Huang et~al.(2020)Huang, Wang, Wei, Huang, Shi, Liu, and
  Huang]{Huang_Wang_Wei_Huang_Shi_Liu_Huang_2020}
Zilong Huang, Xinggang Wang, Yunchao Wei, Lichao Huang, Humphrey Shi, Wenyu
  Liu, and Thomas~S. Huang.
\newblock Ccnet: Criss-cross attention for semantic segmentation.
\newblock \emph{IEEE Transactions on Pattern Analysis and Machine
  Intelligence}, page 1–1, 2020.

\bibitem[Jang and Kim(2019)]{JangWonDong2019InteractiveIS}
Won{-}Dong Jang and Chang{-}Su Kim.
\newblock Interactive image segmentation via backpropagating refinement scheme.
\newblock In \emph{{IEEE} Conference on Computer Vision and Pattern
  Recognition, {CVPR} 2019, Long Beach, CA, USA, June 16-20, 2019}, pages
  5297--5306. Computer Vision Foundation / {IEEE}, 2019.

\bibitem[Kevin~McGuinness(2010)]{KevinMcGuinness2010ACE}
Noel E.~O'Connor Kevin~McGuinness.
\newblock A comparative evaluation of interactive segmentation algorithms.
\newblock \emph{Pattern Recognition}, 2010.

\bibitem[Khan et~al.(2022)Khan, Naseer, Hayat, Zamir, Khan, and
  Shah]{khan2022transformers}
Salman Khan, Muzammal Naseer, Munawar Hayat, Syed~Waqas Zamir, Fahad~Shahbaz
  Khan, and Mubarak Shah.
\newblock Transformers in vision: A survey.
\newblock \emph{ACM computing surveys (CSUR)}, 54\penalty0 (10s):\penalty0
  1--41, 2022.

\bibitem[Kirillov et~al.(2023)Kirillov, Mintun, Ravi, Mao, Rolland, Gustafson,
  Xiao, Whitehead, Berg, Lo, Doll{\'a}r, and Girshick]{kirillov2023segany}
Alexander Kirillov, Eric Mintun, Nikhila Ravi, Hanzi Mao, Chloe Rolland, Laura
  Gustafson, Tete Xiao, Spencer Whitehead, Alexander~C. Berg, Wan-Yen Lo, Piotr
  Doll{\'a}r, and Ross Girshick.
\newblock Segment anything.
\newblock \emph{arXiv:2304.02643}, 2023.

\bibitem[Konstantin~Sofiiuk(2021)]{KonstantinSofiiuk2021RevivingIT}
Anton~Konushin Konstantin~Sofiiuk, Ilia A.~Petrov.
\newblock Reviving iterative training with mask guidance for interactive
  segmentation.
\newblock \emph{arXiv: Computer Vision and Pattern Recognition}, 2021.

\bibitem[Lai et~al.(2023)Lai, Tian, Chen, Li, Yuan, Liu, and Jia]{lai2023lisa}
Xin Lai, Zhuotao Tian, Yukang Chen, Yanwei Li, Yuhui Yuan, Shu Liu, and Jiaya
  Jia.
\newblock Lisa: Reasoning segmentation via large language model.
\newblock \emph{arXiv preprint arXiv:2308.00692}, 2023.

\bibitem[Li et~al.()Li, Mao, Girshick, and He]{YanghaoLiExploringPV}
Yanghao Li, Hanzi Mao, Ross Girshick, and Kaiming He.
\newblock Exploring plain vision transformer backbones for object detection.

\bibitem[Li et~al.(2018)Li, Chen, and Koltun]{ZhuwenLi2018InteractiveIS}
Zhuwen Li, Qifeng Chen, and Vladlen Koltun.
\newblock Interactive image segmentation with latent diversity.
\newblock In \emph{2018 {IEEE} Conference on Computer Vision and Pattern
  Recognition, {CVPR} 2018, Salt Lake City, UT, USA, June 18-22, 2018}, pages
  577--585. {IEEE} Computer Society, 2018.

\bibitem[Lin et~al.(2014)Lin, Maire, Belongie, Hays, Perona, Ramanan,
  Doll{\'a}r, and Zitnick]{TsungYiLin2014MicrosoftCC}
Tsung-Yi Lin, Michael Maire, Serge Belongie, James Hays, Pietro Perona, Deva
  Ramanan, Piotr Doll{\'a}r, and C.~Lawrence Zitnick.
\newblock Microsoft coco: Common objects in context.
\newblock \emph{Lecture Notes in Computer Science}, 2014.

\bibitem[Lin et~al.(2020)Lin, Zhang, Chen, Cheng, and
  Lu]{ZhengLin2020InteractiveIS}
Zheng Lin, Zhao Zhang, Lin{-}Zhuo Chen, Ming{-}Ming Cheng, and Shao{-}Ping Lu.
\newblock Interactive image segmentation with first click attention.
\newblock In \emph{2020 {IEEE/CVF} Conference on Computer Vision and Pattern
  Recognition, {CVPR} 2020, Seattle, WA, USA, June 13-19, 2020}, pages
  13336--13345. {IEEE}, 2020.

\bibitem[Lin et~al.(2022)Lin, Duan, Zhang, Guo, and Cheng]{ZhengLinFocusCutDI}
Zheng Lin, Zheng-Peng Duan, Zhao Zhang, Chun-Le Guo, and Ming-Ming Cheng.
\newblock Focuscut: Diving into a focus view in interactive segmentation.
\newblock pages 2637--2646, 2022.

\bibitem[Liu et~al.(2022{\natexlab{a}})Liu, Xu, Bertasius, and
  Niethammer]{QinLiu2022SimpleClickII}
Qin Liu, Zhenlin Xu, Gedas Bertasius, and Marc Niethammer.
\newblock Simpleclick: Interactive image segmentation with simple vision
  transformers.
\newblock \emph{arXiv preprint arXiv:2210.11006}, 2022{\natexlab{a}}.

\bibitem[Liu et~al.(2022{\natexlab{b}})Liu, Zheng, Planche, Karanam, Chen,
  Niethammer, and Wu]{QinLiu2022PseudoClickII}
Qin Liu, Meng Zheng, Benjamin Planche, Srikrishna Karanam, Terrence Chen, Marc
  Niethammer, and Ziyan Wu.
\newblock Pseudoclick: Interactive image segmentation with click imitation.
\newblock pages 728--745, 2022{\natexlab{b}}.

\bibitem[Liu et~al.(2021)Liu, Lin, Cao, Hu, Wei, Zhang, Lin, and
  Guo]{liu2021swin}
Ze Liu, Yutong Lin, Yue Cao, Han Hu, Yixuan Wei, Zheng Zhang, Stephen Lin, and
  Baining Guo.
\newblock Swin transformer: Hierarchical vision transformer using shifted
  windows.
\newblock In \emph{Proceedings of the IEEE/CVF international conference on
  computer vision}, pages 10012--10022, 2021.

\bibitem[Liu et~al.(2022{\natexlab{c}})Liu, Hu, Lin, Yao, Xie, Wei, Ning, Cao,
  Zhang, Dong, et~al.]{liu2022swin}
Ze Liu, Han Hu, Yutong Lin, Zhuliang Yao, Zhenda Xie, Yixuan Wei, Jia Ning, Yue
  Cao, Zheng Zhang, Li Dong, et~al.
\newblock Swin transformer v2: Scaling up capacity and resolution.
\newblock In \emph{Proceedings of the IEEE/CVF conference on computer vision
  and pattern recognition}, pages 12009--12019, 2022{\natexlab{c}}.

\bibitem[Ma and Wang(2023)]{ma2023segment}
Jun Ma and Bo Wang.
\newblock Segment anything in medical images.
\newblock \emph{arXiv preprint arXiv:2304.12306}, 2023.

\bibitem[Maninis et~al.(2018)Maninis, Caelles, Pont{-}Tuset, and
  Gool]{KevisKokitsiManinis2017DeepEC}
Kevis{-}Kokitsi Maninis, Sergi Caelles, Jordi Pont{-}Tuset, and Luc~Van Gool.
\newblock Deep extreme cut: From extreme points to object segmentation.
\newblock In \emph{2018 {IEEE} Conference on Computer Vision and Pattern
  Recognition, {CVPR} 2018, Salt Lake City, UT, USA, June 18-22, 2018}, pages
  616--625. {IEEE} Computer Society, 2018.

\bibitem[Mazurowski et~al.(2023)Mazurowski, Dong, Gu, Yang, Konz, and
  Zhang]{mazurowski2023segment}
Maciej~A Mazurowski, Haoyu Dong, Hanxue Gu, Jichen Yang, Nicholas Konz, and
  Yixin Zhang.
\newblock Segment anything model for medical image analysis: an experimental
  study.
\newblock \emph{Medical Image Analysis}, 89:\penalty0 102918, 2023.

\bibitem[Perazzi et~al.(2016)Perazzi, Pont{-}Tuset, McWilliams, Gool, Gross,
  and Sorkine{-}Hornung]{FedericoPerazzi2016ABD}
Federico Perazzi, Jordi Pont{-}Tuset, Brian McWilliams, Luc~Van Gool, Markus~H.
  Gross, and Alexander Sorkine{-}Hornung.
\newblock A benchmark dataset and evaluation methodology for video object
  segmentation.
\newblock In \emph{2016 {IEEE} Conference on Computer Vision and Pattern
  Recognition, {CVPR} 2016, Las Vegas, NV, USA, June 27-30, 2016}, pages
  724--732. {IEEE} Computer Society, 2016.

\bibitem[Ren et~al.()Ren, Li, Wang, Xiao, and Chang]{Ren_Li_Wang_Xiao_Chang}
Pengzhen Ren, Changlin Li, Guangrun Wang, Yun Xiao, and
  QingDuXiaodanLiangXiaojun Chang.
\newblock Beyond fixation: Dynamic window visual transformer.

\bibitem[Rother et~al.(2004)Rother, Kolmogorov, and
  Blake]{CarstenRother2004GrabCutIF}
Carsten Rother, Vladimir Kolmogorov, and Andrew Blake.
\newblock "grabcut" interactive foreground extraction using iterated graph
  cuts.
\newblock \emph{ACM transactions on graphics (TOG)}, 23\penalty0 (3):\penalty0
  309--314, 2004.

\bibitem[Sofiiuk et~al.(2020)Sofiiuk, Petrov, Barinova, and
  Konushin]{KonstantinSofiiuk2020fBRSRB}
Konstantin Sofiiuk, Ilia~A. Petrov, Olga Barinova, and Anton Konushin.
\newblock {F-BRS:} rethinking backpropagating refinement for interactive
  segmentation.
\newblock In \emph{2020 {IEEE/CVF} Conference on Computer Vision and Pattern
  Recognition, {CVPR} 2020, Seattle, WA, USA, June 13-19, 2020}, pages
  8620--8629. {IEEE}, 2020.

\bibitem[Strudel et~al.(2021)Strudel, Garcia, Laptev, and
  Schmid]{RobinStrudel2021SegmenterTF}
Robin Strudel, Ricardo Garcia, Ivan Laptev, and Cordelia Schmid.
\newblock Segmenter: Transformer for semantic segmentation.
\newblock pages 7262--7272, 2021.

\bibitem[Vaswani et~al.(2017)Vaswani, Shazeer, Parmar, Uszkoreit, Jones, Gomez,
  Kaiser, and Polosukhin]{AshishVaswani2017AttentionIA}
Ashish Vaswani, Noam Shazeer, Niki Parmar, Jakob Uszkoreit, Llion Jones,
  Aidan~N. Gomez, Lukasz Kaiser, and Illia Polosukhin.
\newblock Attention is all you need.
\newblock In \emph{Advances in Neural Information Processing Systems 30: Annual
  Conference on Neural Information Processing Systems 2017, December 4-9, 2017,
  Long Beach, CA, {USA}}, pages 5998--6008, 2017.

\bibitem[Vaswani et~al.(2021)Vaswani, Ramachandran, Srinivas, Parmar, Hechtman,
  and Shlens]{Vaswani_Ramachandran_Srinivas_Parmar_Hechtman_Shlens_2021}
Ashish Vaswani, Prajit Ramachandran, Aravind Srinivas, Niki Parmar, Blake
  Hechtman, and Jonathon Shlens.
\newblock Scaling local self-attention for parameter efficient visual
  backbones.
\newblock In \emph{2021 IEEE/CVF Conference on Computer Vision and Pattern
  Recognition (CVPR)}, 2021.

\bibitem[Wang et~al.(2023{\natexlab{a}})Wang, Liu, Zhao, Wu, Ma, Yu, Dai, Yang,
  Liu, Zhang, et~al.]{wang2023review}
Jiaqi Wang, Zhengliang Liu, Lin Zhao, Zihao Wu, Chong Ma, Sigang Yu, Haixing
  Dai, Qiushi Yang, Yiheng Liu, Songyao Zhang, et~al.
\newblock Review of large vision models and visual prompt engineering.
\newblock \emph{arXiv preprint arXiv:2307.00855}, 2023{\natexlab{a}}.

\bibitem[Wang et~al.(2021)Wang, Yao, Chen, Cai, He, and
  Liu]{Wang_Yao_Chen_Cai_He_Liu_2021}
Wenxiao Wang, Lu Yao, Long Chen, Deng Cai, Xiaofei He, and Wei Liu.
\newblock Crossformer: A versatile vision transformer based on cross-scale
  attention.
\newblock \emph{arXiv: Computer Vision and Pattern Recognition,arXiv: Computer
  Vision and Pattern Recognition}, 2021.

\bibitem[Wang et~al.(2023{\natexlab{b}})Wang, Dai, Chen, Huang, Li, Zhu, Hu,
  Lu, Lu, Li, et~al.]{wang2023internimage}
Wenhai Wang, Jifeng Dai, Zhe Chen, Zhenhang Huang, Zhiqi Li, Xizhou Zhu,
  Xiaowei Hu, Tong Lu, Lewei Lu, Hongsheng Li, et~al.
\newblock Internimage: Exploring large-scale vision foundation models with
  deformable convolutions.
\newblock In \emph{Proceedings of the IEEE/CVF Conference on Computer Vision
  and Pattern Recognition}, pages 14408--14419, 2023{\natexlab{b}}.

\bibitem[Wu et~al.(2023)Wu, Fu, Fang, Liu, Wang, Xu, Jin, and
  Arbel]{wu2023medical}
Junde Wu, Rao Fu, Huihui Fang, Yuanpei Liu, Zhaowei Wang, Yanwu Xu, Yueming
  Jin, and Tal Arbel.
\newblock Medical sam adapter: Adapting segment anything model for medical
  image segmentation.
\newblock \emph{arXiv preprint arXiv:2304.12620}, 2023.

\bibitem[Xie et~al.(2021)Xie, Wang, Yu, Anandkumar, Alvarez, and
  Luo]{Xie_Wang_Yu_Anandkumar_Alvarez_Luo_2021}
Enze Xie, Wenhai Wang, Zhiding Yu, Animashree Anandkumar, JoseM. Alvarez, and
  Ping Luo.
\newblock Segformer: Simple and efficient design for semantic segmentation with
  transformers.
\newblock \emph{Cornell University - arXiv,Cornell University - arXiv}, 2021.

\bibitem[Xu et~al.(2016)Xu, Price, Cohen, Yang, and Huang]{NingXu2016DeepIO}
Ning Xu, Brian~L. Price, Scott Cohen, Jimei Yang, and Thomas~S. Huang.
\newblock Deep interactive object selection.
\newblock In \emph{2016 {IEEE} Conference on Computer Vision and Pattern
  Recognition, {CVPR} 2016, Las Vegas, NV, USA, June 27-30, 2016}, pages
  373--381. {IEEE} Computer Society, 2016.

\bibitem[Yang et~al.()Yang, Li, Zhang, Dai, Xiao, Yuan, Gao, Redmond, Cloud,
  and Ai]{Yang_Li_Zhang_Dai_Xiao_Yuan_Gao_Redmond_Cloud_Ai}
Jianwei Yang, Chunyuan Li, Pengchuan Zhang, Xiyang Dai, Bin Xiao, Lu Yuan,
  Jianfeng Gao, MicrosoftResearchAt Redmond, Microsoft Cloud, and + Ai.
\newblock Focal self-attention for local-global interactions in vision
  transformers.

\bibitem[Yu et~al.(2023)Yu, Feng, Feng, Liu, Jin, Zeng, and
  Chen]{yu2023inpaint}
Tao Yu, Runseng Feng, Ruoyu Feng, Jinming Liu, Xin Jin, Wenjun Zeng, and Zhibo
  Chen.
\newblock Inpaint anything: Segment anything meets image inpainting.
\newblock \emph{arXiv preprint arXiv:2304.06790}, 2023.

\bibitem[Yuan et~al.(2021)Yuan, Fu, Huang, Lin, Zhang, Chen, and
  Wang]{yuan2021hrformer}
Yuhui Yuan, Rao Fu, Lang Huang, Weihong Lin, Chao Zhang, Xilin Chen, and
  Jingdong Wang.
\newblock Hrformer: High-resolution vision transformer for dense predict.
\newblock \emph{Advances in Neural Information Processing Systems},
  34:\penalty0 7281--7293, 2021.

\bibitem[Zhang et~al.(2020)Zhang, Liew, Wei, Wei, and
  Zhao]{ShiyinZhang2020InteractiveOS}
Shiyin Zhang, Jun~Hao Liew, Yunchao Wei, Shikui Wei, and Yao Zhao.
\newblock Interactive object segmentation with inside-outside guidance.
\newblock In \emph{2020 {IEEE/CVF} Conference on Computer Vision and Pattern
  Recognition, {CVPR} 2020, Seattle, WA, USA, June 13-19, 2020}, pages
  12231--12241. {IEEE}, 2020.

\bibitem[Zhuoran et~al.(2021)Zhuoran, Mingyuan, Haiyu, Shuai, and
  Hongsheng]{Zhuoran_Mingyuan_Haiyu_Shuai_Hongsheng_2021}
Shen Zhuoran, Zhang Mingyuan, Zhao Haiyu, Yi Shuai, and Li Hongsheng.
\newblock Efficient attention: Attention with linear complexities.
\newblock In \emph{2021 IEEE Winter Conference on Applications of Computer
  Vision (WACV)}, 2021.

\end{thebibliography}
}

\clearpage
\setcounter{page}{1}
\maketitlesupplementary


\appendix

\noindent The "SAM vs FocSAM.mp4" comparison video is available in the supplemental materials.

\section{Implementation Details}

\subsection{Datasets}

In this study, we conduct experiments on six datasets to assess our methods comprehensively:
\begin{itemize}
    \item \textbf{GrabCut}~\cite{CarstenRother2004GrabCutIF}: Features 50 images, each with distinct foreground and background, totaling 50 instances.
    \item \textbf{Berkeley}~\cite{KevinMcGuinness2010ACE}: Comprises 96 images (100 instances), with some overlap with GrabCut.
    \item \textbf{DAVIS}~\cite{FedericoPerazzi2016ABD}: Focuses on 345 specific frames from 50 videos, aligning with previous studies~\cite{KonstantinSofiiuk2021RevivingIT,XiChen2022FocalClickTP,QinLiu2022SimpleClickII}.
    \item \textbf{SBD}~\cite{BharathHariharan2011SemanticCF}: Includes 2857 validation images with 6671 instances for evaluation purposes.
    \item \textbf{MVTec}~\cite{bergmann2019mvtec}: Selected for its high-quality pixel-wise annotations of industrial defects, ideal for interactive segmentation's practical applications. Specific defects like cut-lead and misplaced elements in transistors are excluded due to their misalignment with image segmentation, refining the dataset to 1238 instances.
    \item \textbf{COD10K}~\cite{fan2020camouflaged}: Contains 2026 instances of camouflaged objects that blend into their backgrounds, providing a distinct challenge for interactive segmentation.
\end{itemize}

\subsection{Training Details}

In training our proposed FocSAM on COCO~\cite{TsungYiLin2014MicrosoftCC} and LVIS~\cite{AgrimGupta2019LVISAD}, we adopt the AdamW optimizer~\cite{loshchilov2017decoupled}. The initial learning rate is set to \(1e-6\) for the first \(1,500\) iterations, which is then raised to \(1e-4\). We then apply a polynomial decay to the learning rate, setting AdamW's \(\beta_1\) to \(0.9\) and \(\beta_2\) to \(0.999\). Our batch size is \(4\) per GPU, totaling \(16\) samples across \(4\) GPUs, and images are resized and padded to \(1024 \times 1024\). We attempt to jointly train FocSAM with the SAM decoder, but such a strategy results in unstable training. Therefore, we fine-tune the SAM decoder~\cite{kirillov2023segany} alone over \(320,000\) iterations at the first stage. Then, at the second stage we freeze the trained decoder and train the FocSAM's focus refiner for additional \(160,000\) iterations.

\subsection{Click Simulation}

During training, we adopt the click simulation strategy from InterFormer~\cite{huang2023interformer} due to its simplicity. We set the upper limit for simulated clicks at 20. To determine the distribution of click counts, we employ a decay coefficient $\gamma$, where the probability for a given number of clicks decreases progressively. Specifically, the probability of having $i$ clicks is $\gamma$ times the probability of having $i-1$ clicks, with the constraint that $\gamma < 1$. This method ensures a higher likelihood of selecting fewer clicks, reducing computational costs. For joint training on COCO~\cite{TsungYiLin2014MicrosoftCC} and LVIS~\cite{AgrimGupta2019LVISAD} datasets, InterFormer~\cite{huang2023interformer} sets \(\gamma\) at 0.6 for both. Instead, to avoid bias towards small objects in LVIS, we use different \(\gamma\) values for COCO (\(\gamma = 0.6\)) and LVIS (\(\gamma = 0.9\)). This adjustment allows for more effective use of LVIS's detailed annotations in the later refinement stages. In FocSAM, we first decide the number of clicks, $N$, and then determine the refinement step, $K$, using a similar sampling strategy, where we set distinct $\gamma_r$ values for COCO ($\gamma_r = 0.6$) and LVIS ($\gamma_r = 0.35$) with $N$ as the upper limit to ensure a similar refinement process. After determining the $N$ and $K$ (only for FocSAM), SAM and FocSAM perform click simulations on training images using GT as an oracle to specify clicks randomly within incorrectly predicted regions. 

\section{Ablation Study}

\subsection{Convergence Analysis}

\begin{figure*}[ht]
    \centering
    \includegraphics[width=1.0\linewidth]{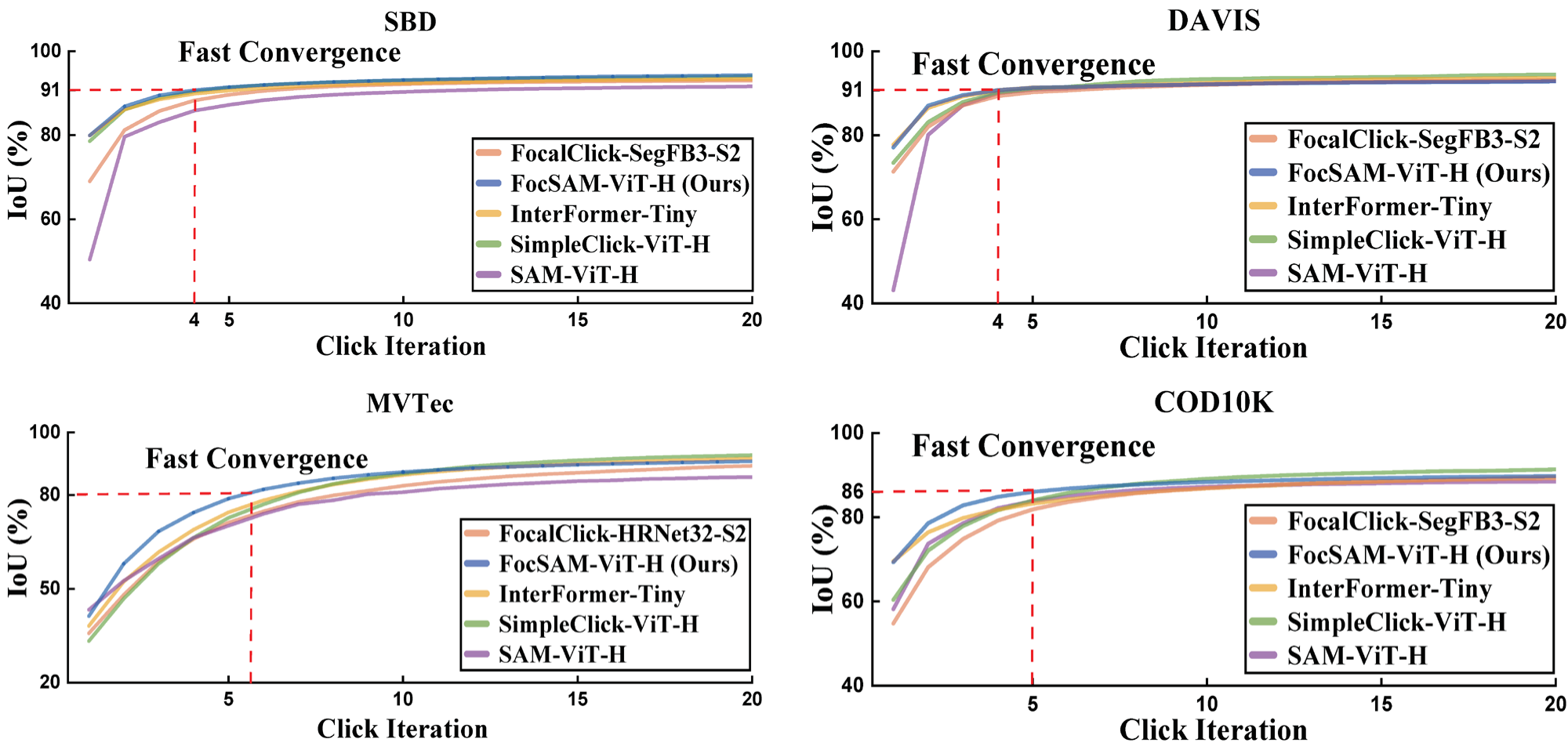}
    \vspace{-1.5em}
    \caption{Convergence Analysis. Each subfigure displays the average IoU for all samples at successive clicks. These plots illustrate the rapid convergence of FocSAM, which achieves high IoU values with only a few clicks.}
    \vspace{-0.5em}
    \label{fig:convergence}
\end{figure*}

\begin{table*}[ht]
    \centering
    \begin{tabular}{@{}lcccccc@{}}
        \toprule
        \multicolumn{1}{c}{\multirow{2}{*}{\textbf{Method}}} & \multicolumn{2}{c}{\textbf{SBD}} & \multicolumn{2}{c}{\textbf{MVTec}} & \multicolumn{2}{c}{\textbf{COD10K}} \\
        \cmidrule(r){2-3} \cmidrule(lr){4-5} \cmidrule(l){6-7}
        & 20NoC@90 & 100NoC@95 & 20NoC@90 & 100NoC@95 & 20NoC@90 & 100NoC@95 \\
        \midrule
        SAM (w/o BBox) & 7.62 & 63.40 & 13.97 & 81.90 & 10.36 & 76.73 \\
        SAM (w/ BBox)  & 7.27 & 63.28 & 13.71 & 82.08 & 10.66 & 76.65 \\
        FocSAM & 4.69 & 32.96 & 11.14 & 62.82 & 8.91 & 62.61 \\
        \bottomrule
    \end{tabular}
    \vspace{-0.5em}
    \caption{Ablation study on SAM with bounding boxes.}
    \vspace{-0.5em}
    \label{table:bbox}
\end{table*}

\begin{table*}[!ht]
    \centering
    \begin{tabular}{@{}lcccccc@{}}
        \toprule
        \multicolumn{1}{c}{\multirow{2}{*}{\textbf{Dwin-MSA}}} & \multicolumn{2}{c}{\textbf{SBD}} & \multicolumn{2}{c}{\textbf{MVTec}} & \multicolumn{2}{c}{\textbf{COD10K}} \\
        \cmidrule(r){2-3} \cmidrule(lr){4-5} \cmidrule(l){6-7}
        & 20NoC@90 & 100NoC@95 & 20NoC@90 & 100NoC@95 & 20NoC@90 & 100NoC@95 \\
        \midrule
        Window-16 & 4.69 & 32.96 & 11.14 & 62.82 & 8.91 & 62.61 \\
        Window-8  & 4.75 & 34.08 & 11.31 & 64.69 & 9.21 & 64.17 \\
        Window-32 & 4.85 & 33.95 & 11.27 & 63.54 & 9.25 & 64.18 \\
        \bottomrule
    \end{tabular}
    \vspace{-0.5em}
    \caption{Ablation study on Dwin-MSA's window sizes. 
    }
    \vspace{-1em}
    \label{tab:dwinwindow}
\end{table*}


We perform convergence analysis experiments on the SBD~\cite{BharathHariharan2011SemanticCF}, DAVIS~\cite{FedericoPerazzi2016ABD}, MVTec~\cite{bergmann2019mvtec}, and COD10K~\cite{fan2020camouflaged} datasets with sufficient samples. In these experiments, we compute the average IoU for all samples at each click, comparing our FocSAM with previous methods~\cite{XiChen2022FocalClickTP,QinLiu2022SimpleClickII,huang2023interformer}. As depicted in Figure~\ref{fig:convergence}, the results showcase FocSAM's fast convergence across these datasets. FocSAM notably achieves high IoU values with only a few clicks. Such rapid convergence is particularly pronounced in the challenging MVTec~\cite{bergmann2019mvtec} and COD10K~\cite{fan2020camouflaged} datasets, where FocSAM outperforms other methods, including the previous state-of-the-art SimpleClick-ViT-H~\cite{QinLiu2022SimpleClickII}. In SBD~\cite{BharathHariharan2011SemanticCF} and DAVIS~\cite{FedericoPerazzi2016ABD} datasets, FocSAM demonstrates a convergence rate on par with SimpleClick-ViT-H~\cite{QinLiu2022SimpleClickII}, underscoring its efficiency in various interactive segmentation scenarios.

\subsection{SAM's Bounding Box Prompt}

\noindent\textbf{Experimental settings.}
SAM~\cite{kirillov2023segany} can simultaneously process click and bounding box prompts. Notably, in our proposed FocSAM, the Dwin-MSA module conceptually shares similarities with the processing of bounding box prompts. Therefore, we evaluate SAM with additional bounding boxes around target objects for ablation studies. Specifically, we utilize the GT to find the bounding box encompassing the target object and expand it by $1.4\times$ to include the context of the surrounding area. During the interactive segmentation of SAM, these boxes are supplied as an additional prompt. Likewise, we report the results on SBD~\cite{BharathHariharan2011SemanticCF}, MVTec~\cite{bergmann2019mvtec}, and COD10K~\cite{fan2020camouflaged} datasets, including the metrics 20NoC@90 and 100NoC@95.

\noindent\textbf{Results.}
Table~\ref{table:bbox} reveals that integrating interactive information from bounding boxes offers marginal improvement to SAM's performance. This demonstrates that SAM cannot fully exploit the potential of such interactive information from the additional boxes. 
In contrast, FocSAM effectively utilizes similar information through its Dwin-MSA module. Specifically, FocSAM enhances the performance by overlaying bounding boxes on previous predictions and feeding these into the Dwin-MSA module to select windows relevant to the object. This approach underscores FocSAM's efficiency in leveraging available information for enhanced performance.

\subsection{Impact of Dwin-MSA's Window Size}
In Table~\ref{tab:dwinwindow}, our ablation study on Dwin-MSA's window size indicates window-16 outperforms both window-8 and window-32. The limited attention scope of window-8 constrains its performance. In contrast, while window-32 has a broader attention span, it incorporates excessive object-unrelated areas, which undermines its effectiveness.

\section{Qualitative Results}

\begin{figure*}
    \centering
    \includegraphics[width=0.84\linewidth]{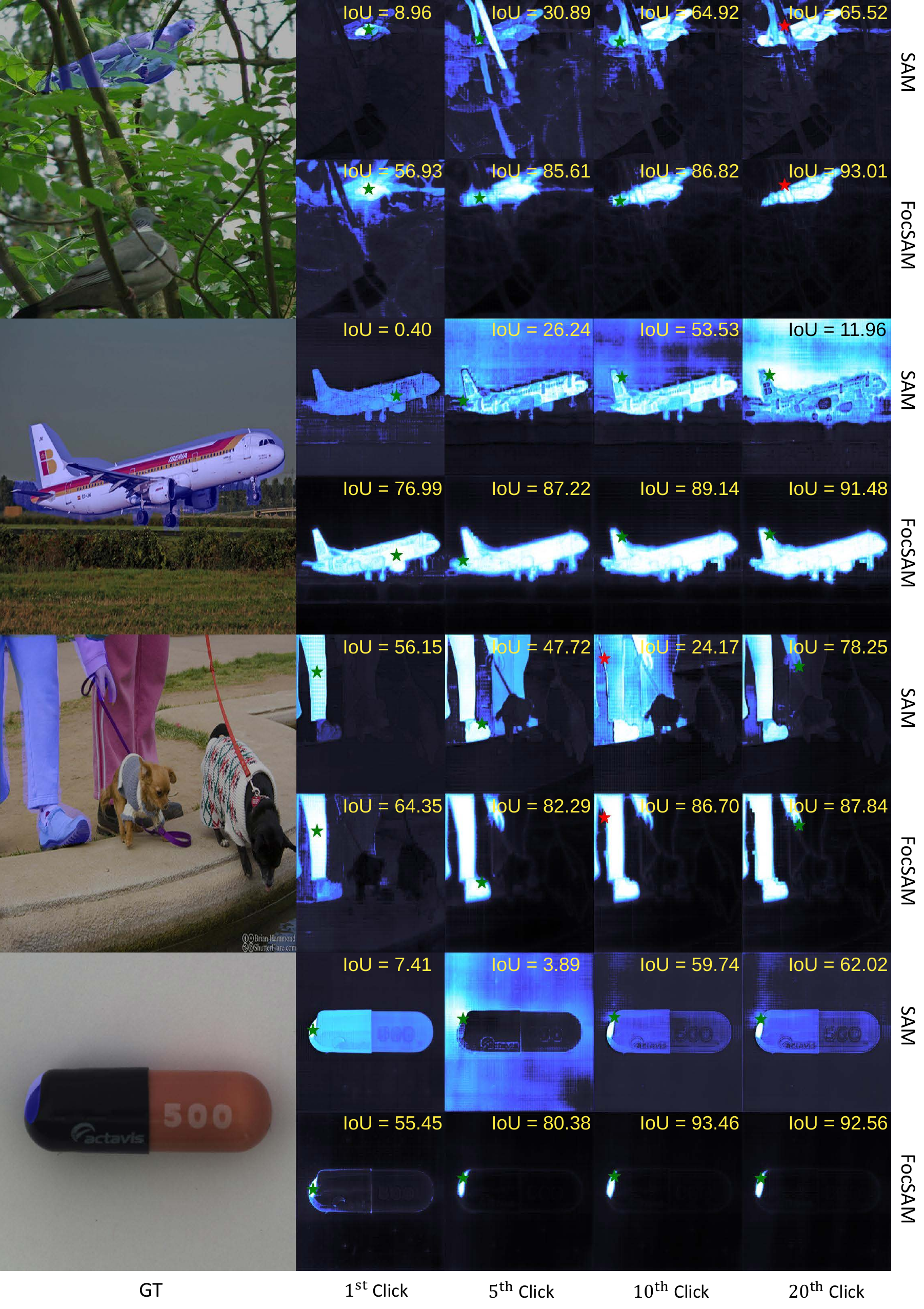}
    \vspace{-1em}
    \caption{Qualitative results (1). On the left, an example is depicted with an image overlaid with its GT (blue mask). To the right, two rows display interactive segmentation results at the $1$st, $5$th, $10$th, and $20$th clicks, where the most recent click is highlighted with a star, green for positive and red for negative feedback. The top row illustrates the results from SAM, and the bottom row shows those from FocSAM. These visual comparisons reveal the segmentation efficiency of FocSAM and SAM at different stages of annotator clicks.}
    \label{fig:qualitative}
\end{figure*}

\begin{figure*}
    \centering
    \includegraphics[width=0.84\linewidth]{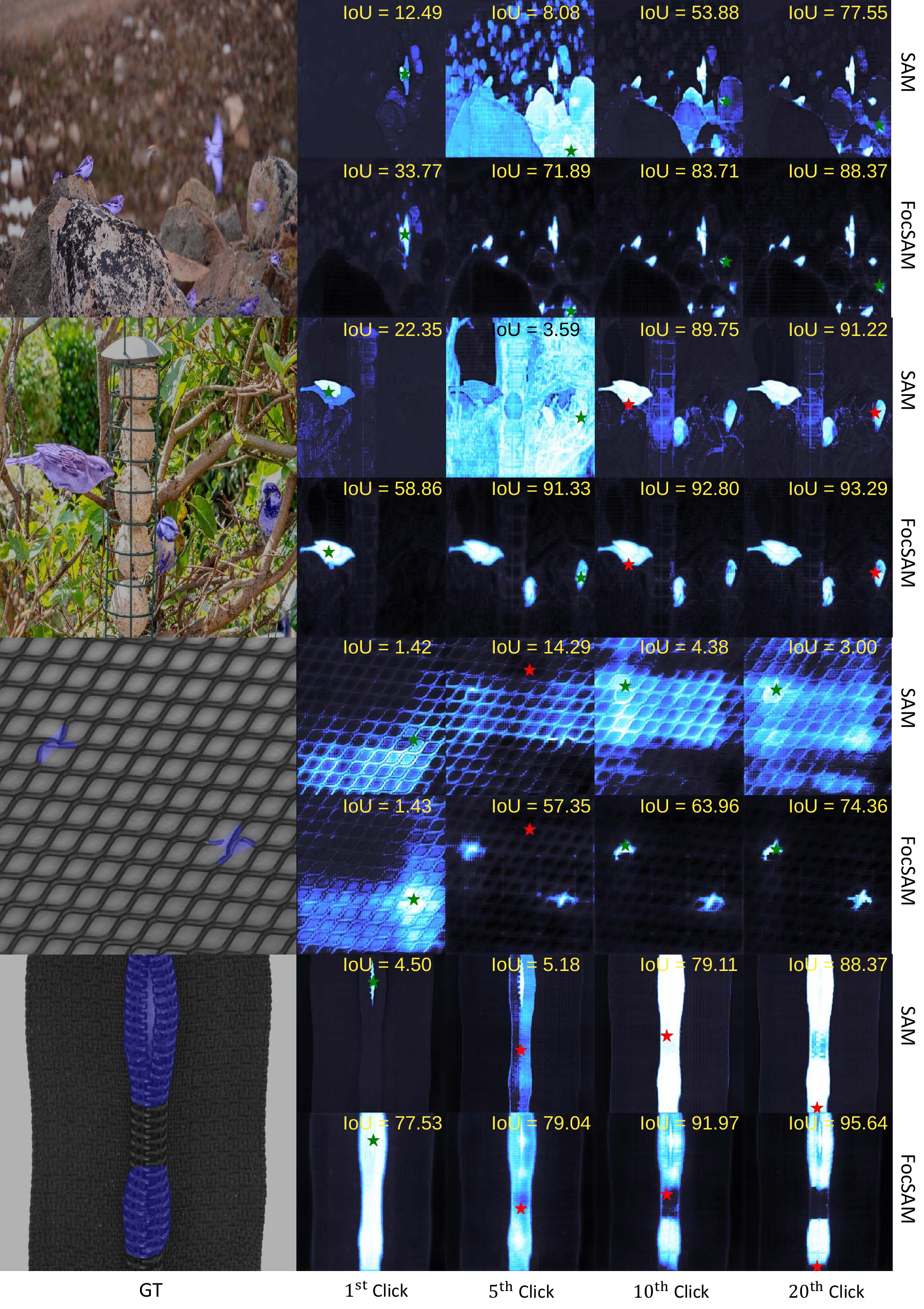}
    \vspace{-1em}
    \caption{Qualitative results (2). On the left, an example is depicted with an image overlaid with its GT (blue mask). To the right, two rows display interactive segmentation results at the $1$st, $5$th, $10$th, and $20$th clicks, where the most recent click is highlighted with a star, green for positive and red for negative feedback. The top row illustrates the results from SAM, and the bottom row shows those from FocSAM. These visual comparisons reveal the segmentation efficiency of FocSAM and SAM at different stages of annotator clicks.}
    \label{fig:qualitative-2}
\end{figure*}

In Figure~\ref{fig:qualitative}~\ref{fig:qualitative-2}, we present the interactive segmentation results of FocSAM and SAM across various scenarios. For a more comprehensive set of results, please refer to the accompanying video titled ``SAM vs FocSAM.mp4.''

\end{document}